\documentclass[runningheads]{llncs}

\usepackage{eccv}

\usepackage{eccvabbrv}

\usepackage{graphicx}
\usepackage{booktabs}

\usepackage{multirow}
\usepackage{colortbl}
\usepackage{xcolor}
\definecolor{best}{rgb}{1,0,0} % red color for best
\definecolor{second}{rgb}{0,0,1} % blue color for best
\usepackage{arydshln}
\usepackage{wrapfig}
\usepackage{subcaption}
\usepackage[accsupp]{axessibility}  % Improves PDF readability for those with disabilities.

\usepackage{hyperref}

\usepackage{orcidlink}

\AtBeginDocument{%
  \setlength{\abovedisplayskip}{5pt plus 1pt minus 2pt}%
  \setlength{\belowdisplayskip}{5pt plus 1pt minus 2pt}%
  \setlength{\abovedisplayshortskip}{3pt plus 1pt minus 1pt}%
  \setlength{\belowdisplayshortskip}{3pt plus 1pt minus 1pt}%
}

\begin{document}

% ---------------------------------------------------------------
\title{FreqOrtho-SR: Frequency-Guided Orthogonal Expert Learning for Real-World Image Super-Resolution} 

\titlerunning{FreqOrtho-SR}

\author{Minh Son Hoang\thanks{Equal contribution.} \and
Dinh Phu Tran\textsuperscript{$\star$} \and
Quyen Nguyen Duc \and
Dam Hoang Phuong \and
Daeyoung Kim\thanks{Corresponding author.}}

\authorrunning{M.~S.~Hoang, D.~P.~Tran et al.}

\institute{School of Computing, KAIST, Republic of Korea\\
\email{\{sonhm2910,phutx2000,nguyenducquyen2311,hoangphuong1211,kimd\}@kaist.ac.kr}}

\maketitle

\begin{abstract}
Diffusion prior-based methods have shown impressive results in real-world image super-resolution (ISR), yet two key challenges persist: balancing pixel-level fidelity with semantic quality, and adapting to diverse degradations. Existing dual-branch approaches freeze the pixel module during semantic training, but the semantic branch can still expand capacity within the pixel subspace, precluding genuine perceptual improvement. Moreover, using a single static adapter cannot generalize across heterogeneous real-world corruptions.
To address both issues, we propose FreqOrtho-SR, which comprises: \textbf{Freq}uency-guided Mixture of LoRA Experts (FreqMoE), it routes inputs to specialized experts via a non-parametric FFT-based degradation-feature extractor that encodes frequency-domain signatures, enabling stable and interpretable specialization across corruption types; and \textbf{Ortho}gonal Gradient Projection (OGP), which reframes the dual-objective optimization as a subspace-constrained problem: by extracting the pixel-fidelity subspace via SVD on combined expert weight deltas and projecting semantic gradients onto its null space, OGP guarantees orthogonality between the two objectives, enabling genuinely complementary learning without mutual interference. Experiments show that FreqOrtho-SR achieves competitive overall performance and a strong fidelity-perception trade-off across multiple benchmarks with efficient single-step inference.
The source code of our method can be found at \href{https://github.com/sonhm3029/FreqOrtho-SR}{\texttt{sonhm3029/FreqOrtho-SR}}.
  
  \keywords{Image Super-Resolution \and Frequency-guided Mixture of Experts \and Orthogonal Learning \and Diffusion Models}
\end{abstract}

\begin{figure}[t]
  \centering
  \includegraphics[width=0.9\textwidth]{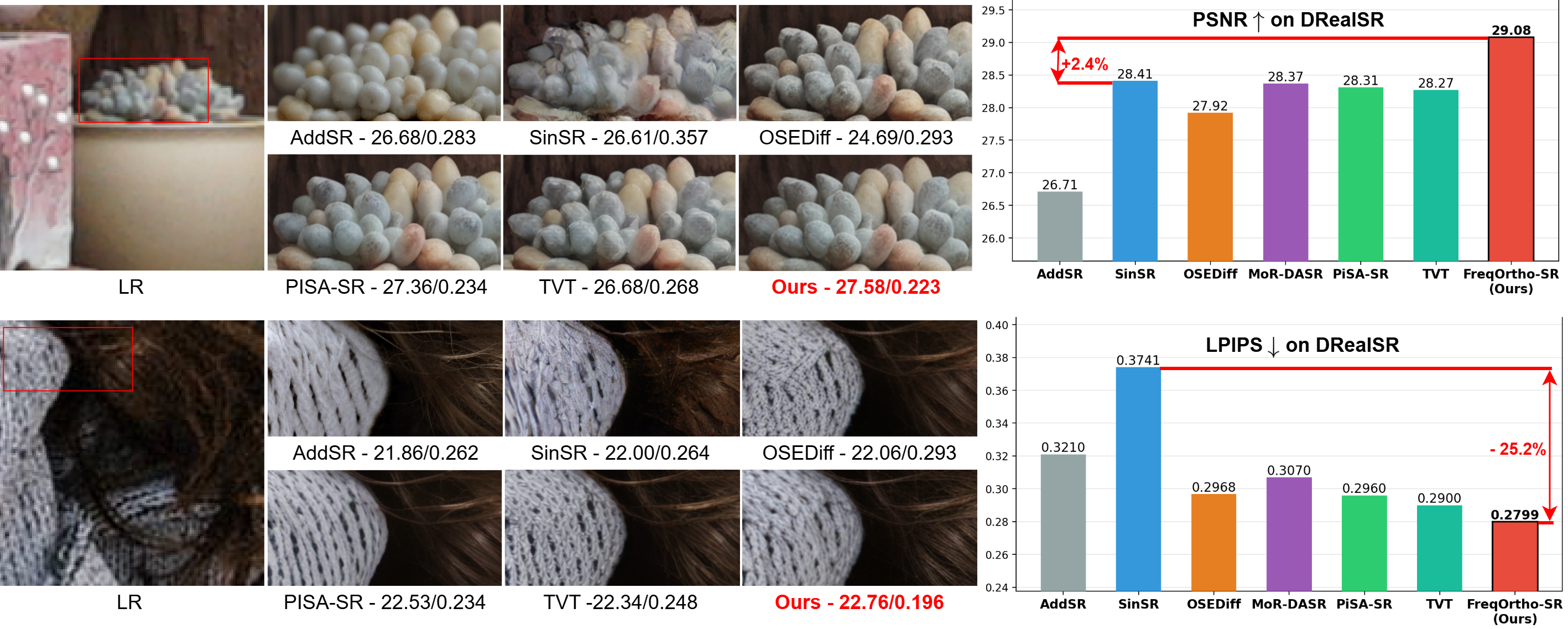}
  \caption{Visual and quantitative comparison on DRealSR. Values below each patch denote PSNR$\uparrow$/LPIPS$\downarrow$, with best in \textcolor{red}{red}. Our FreqOrtho-SR achieves the best fidelity and perceptual quality among one-step diffusion methods.}
  \label{fig:teaser}
\end{figure}

\section{Introduction}
\label{sec:intro}

Image super-resolution (ISR) \cite{wang2020deep, tran2026sat} aims to produce high-resolution (HR) images from degraded low-resolution (LR) inputs. A main challenge in ISR is the inherent trade-off between perception and distortion \cite{blau2018perception}. While pixel-level regression methods using $\mathcal{L}_1$ or $\mathcal{L}_2$ losses achieve high fidelity metrics (\eg, PSNR, SSIM), their results are perceptually unsatisfying and overly smooth \cite{ledig2017photo}. Differently, generative methods create visually appealing textures but also introduce artifacts that compromise content fidelity \cite{liang2022details}. The challenge becomes even more complex in real-world ISR, which must handle diverse, unknown degradations 
% such as motion blur, JPEG compression artifacts, and downsampling distortion 
\cite{zhang2021designing, wang2021real}.

Early methods \cite{dong2015image, kim2016accurate, zhang2018image, dai2019second, kong2021classsr, tran2024channel} employed deep neural networks trained with pixel-level losses, but struggled with real-world degradations, producing outputs that lacked perceptual realism.
GAN-based approaches \cite{ledig2017photo, wang2018esrgan, wang2021real, liang2022efficient, liang2022details} significantly enhanced visual realism by aligning outputs with natural image distributions. However, their limited generative capacity and training instability often lead to unnatural artifacts.
Recent text-to-image diffusion models, particularly Stable Diffusion (SD), have transformed real-world ISR \cite{lin2024diffbir, wang2024exploiting, wu2024seesr, yang2024pixel, yu2024scaling, sun2025pixel, yi2025fine} by providing powerful generative prior knowledge from vast image datasets and yielding more realistic SR outputs than GAN-based methods. However, the iterative sampling process inherent to dynamic models typically requires multiple sequential steps, imposing significant latency that hinders practical deployment.

Recent efforts \cite{wang2024sinsr, wu2024one, sun2025pixel, yi2025fine, he2025mixture} address these limitations via one-step diffusion models, achieved either by distilling multi-step diffusion models or fine-tuning pre-trained diffusion models with Low-Rank Adaptation (LoRA).
OSEDiff \cite{wu2024one} improves the process by inputting LR latent features and uses Variational Score Distillation to condense multi-step diffusion into a single step via LoRA fine-tuning.  
TVT \cite{yi2025fine} addresses fine-structure preservation by transferring 8$\times$ downsampled VAE of SD to a variant that supports 4$\times$ downsampling. PiSA-SR \cite{sun2025pixel} pushes the field forward by using a dual-LoRA framework that decouples pixel-level regression
%(using $\mathcal{L}_2$ loss)
from semantic enhancement,
%(using LPIPS and Classifier Score Distillation losses).
better improving both objectives.

However, these methods still face two main challenges. 
First, many monolithic architectures apply a uniform balance of fidelity and perceptual quality across different degradation types,
%%%%%%%%%%%%% Consider for related work 
failing to adapt to complex scenarios where conflicting restoration needs exist.
Second, jointly optimizing pixel-level fidelity and semantic-level enhancement tends to entangle the two objectives, making it difficult to balance faithful structure preservation and perceptual detail synthesis. PiSA-SR \cite{sun2025pixel} addresses this issue by decoupling them into two LoRA modules and freezing the pixel branch during semantic training. However, we argue that this passive decoupling is still insufficient because it does not explicitly control the geometry of semantic updates. Even with the pixel weights frozen, the semantic branch can receive update components along the pixel-fidelity subspace and learn features already covered by the pixel branch, leading to redundant subspace overlap. This redundancy hampers the model's ability to fully utilize its capacity for genuine perceptual enhancement.

To mitigate these limitations, we propose FreqOrtho-SR, a novel framework for real-world SR with two key innovations:
%%%%%%%%%%%%%%%% Moving to Motivation section %%%%%%%%%%%%%%%%
1) \textit{Frequency-Guided Mixture of Experts} (FreqMoE) employs multiple LoRA experts tailored to specific degradation patterns. Using a lightweight Fast Fourier Transform (FFT) based \cite{cooley1965algorithm} degradation feature extractor to guide a gating network that routes inputs to appropriate experts, enabling stable and interpretable specialization. 
2) \textit{Orthogonal Gradient Projection} (OGP) to mitigate pixel-subspace gradient interference in multi-objective optimization. Inspired by continual learning techniques, we view fidelity preservation and perceptual enhancement as sequential tasks requiring orthogonality. By projecting away semantic-gradient components that lie in the pixel-level subspace, OGP enables the semantic LoRA to learn in an independent subspace, resulting in significantly improved perceptual quality with minimal fidelity trade-off.

Our main contributions can be summarized as follows:

\begin{enumerate}
    \item We propose FreqOrtho-SR, a novel method for degradation-adaptive optimization that introduces a frequency-aware gating mechanism based on explicit FFT-based degradation signatures. This enables stable and interpretable expert specialization, where different experts operate at distinct fidelity-perceptual levels tailored to specific degradation characteristics.
    \item We introduce orthogonal gradient projection from continual learning to real-world ISR task, reducing overlap between pixel and semantic subspaces and enabling the semantic LoRA to learn more effectively in an independent subspace.
    \item Our extensive experiments show that FreqOrtho-SR achieves competitive overall performance, with best or second-best results on several key image quality metrics across multiple datasets.
\end{enumerate}

\section{Related Work}
\label{sec:related_work}

\subsection{Diffusion Model-based Super-Resolution}

Latent Diffusion Models (LDMs) \cite{rombach2022high} extend DDPMs \cite{ho2020denoising} by performing denoising in a compressed latent space, yielding the Stable Diffusion (SD) model whose generative priors are now widely used for real-world SR.

Multi-step methods employ these priors at 20--50 denoising steps. StableSR \cite{wang2024exploiting} injects LR features into a frozen SD model via spatial feature transform layers; DiffBIR \cite{lin2024diffbir} first removes degradations and then enhances details with ControlNet \cite{zhang2023adding}; PASD \cite{yang2024pixel} adds pixel-aware cross-attention for structure guidance; and SeeSR \cite{wu2024seesr} steers generation with degradation-aware text prompts. Despite strong visual quality, iterative sampling remains a practical bottleneck.
One-step alternatives address this cost via consistency distillation (SinSR \cite{wang2024sinsr}), residual-shifting Markov chains (ResShift \cite{yue2024resshift}), Variational Score Distillation with LoRA \cite{hu2022lora} tuning (OSEDiff \cite{wu2024one}), adversarial diffusion distillation (AddSR \cite{tai2026addsr}), target score distillation (TSD-SR \cite{dong2025tsd}), diffusion-inversion noise prediction (InvSR \cite{yue2025invsr}), VAE transfer from $8{\times}$ to $4{\times}$ downsampling (TVT \cite{yi2025fine}), dual-LoRA decoupling of pixel regression and semantic enhancement (PiSA-SR \cite{sun2025pixel}), and CLIP-guided mixture-of-ranks routing (MoR-DASR \cite{he2025mixture}). Unlike MoR-DASR, which routes via a frozen CLIP encoder with predefined prompt pairs, our FreqMoE employs non-parametric FFT-based degradation signatures for interpretable routing without external pretrained models.

However, all these methods face a fundamental contradiction between pixel-level fidelity and semantic-level enhancement. Monolithic architectures (\eg, TVT \cite{yi2025fine}, PiSA-SR \cite{sun2025pixel}, OSEDiff \cite{wu2024one}) lack the flexibility to adapt to diverse real-world degradations, while dual-branch designs (\eg, PiSA-SR \cite{sun2025pixel}) can suffer from subspace overlap between pixel- and semantic-level objectives.
FreqOrtho-SR addresses both issues via frequency-guided expert specialization and orthogonal constraint theory, transforming the dual-branch architecture from a rigid structure into a dynamic, mathematically-constrained framework.

\subsection{Frequency Analysis in Image Processing}

Frequency domain analysis has long been a fundamental aspect of image processing \cite{pratt2007digital, tran2024channel, chen2023swinfsr, rao2021global, li2023discrete, tran2025vsrm}. Our insight is that different types of degradation exhibit unique frequency patterns. For example, blur suppresses high frequencies \cite{kundur1996blind, narvekar2011no}, Gaussian noise elevates all frequencies uniformly \cite{buades2005review, foi2007pointwise}, JPEG compression introduces artifacts in 8$\times$8 blocks that are visible in the frequency spectrum \cite{luo2010jpeg}, and downsampling creates distinct aliasing patterns \cite{blu2004linear}. These observations have been utilized for various tasks such as blur detection \cite{ferzli2009no, vu2011bf}, noise estimation \cite{liu2013single, pyatykh2012image}, and quality assessment \cite{mittal2012making, sheikh2006image}. 
Recent methods have incorporated frequency information through spectral losses \cite{jiang2021focal} and frequency-domain adversarial training \cite{durall2020watch}. However, frequency features have not been systematically applied to routing in mixture of experts (MoE) architectures, which remains a significant challenge for effective routing. Traditional gating based on spatial features may suffer from training instability and load imbalance \cite{shazeer2017outrageously, fedus2022switch}, leading some experts to dominate routing while others remain underutilized. Additionally, learned spatial representations may struggle to capture underlying degradation patterns without explicit structural priors \cite{zhou2022mixture, puigcerver2023sparse}. 
We address this gap by integrating explicit frequency-based degradation features into the gating mechanism.

\subsection{Continual Learning and Gradient Projection}

Continual learning aims to acquire sequential tasks without forgetting previously acquired knowledge \cite{mccloskey1989catastrophic, french1999catastrophic}. One prominent method is Gradient Projection Memory (GPM) \cite{saha2021gradient}, which preserves task-specific knowledge by constraining gradient updates via subspace projection. After learning task A, GPM extracts important parameter directions using Singular Value Decomposition (SVD) \cite{saha2021gradient} and projects the gradients of subsequent tasks into the null space of this subspace to prevent interference \cite{zeng2019continual, chaudhry2019continual}. 
Related methods include Orthogonal Weights Modification \cite{zeng2019continual} and Gradient Episodic Memory \cite{lopez2017gradient}, both of which constrain gradient updates to preserve prior knowledge. While these methods are widely used in classification \cite{chaudhry2019continual, saha2021gradient} and reinforcement learning \cite{wolczyk2021continual}, they have not yet been explored in ISR.
Therefore, we propose orthogonal projection to real-world ISR, treating pixel-level fidelity and semantic-level enhancement as sequential learning tasks. By extracting pixel-level LoRA subspaces via SVD and projecting semantic gradients into the null space, we remove pixel-subspace gradient components that would otherwise drive redundant semantic updates. This enables semantic LoRA to learn in an independent subspace, thereby improving perceptual quality while maintaining competitive fidelity.

\section{Methodology}
\label{sec:methodology}

This section first formulates the SD-based SR as a residual learning model. It then introduces our proposed FreqOrtho-SR, a novel framework designed for true subspace disentanglement and adaptive restoration in real-world ISR, aiming to achieve a better balance between fidelity and perceptual quality. In this study, we denote the low-resolution (LR) and high-resolution (HR) images as \(x_L\) and \(x_H\). Their corresponding latent codes are represented as \(z_L = \mathcal{E}(x_L)\) and \(z_H = \mathcal{E}(x_H)\). We can approximate that \(x_L \approx \mathcal{D}(z_L)\) and \(x_H \approx \mathcal{D}(z_H)\), where \(\mathcal{E}\) and \(\mathcal{D}\) are the encoder and decoder of a pre-trained Variational Autoencoder (VAE).

\begin{figure*}[tb]
  \centering
  \includegraphics[width=\linewidth]{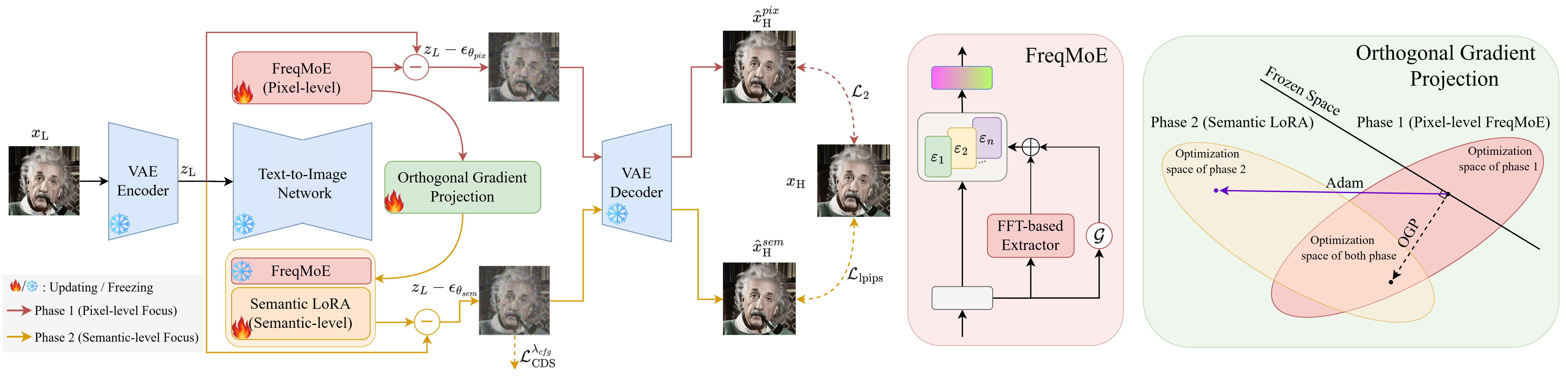}
  \caption{Overall architecture of our FreqOrtho-SR. The FreqMoE and Semantic LoRA modules are optimized for pixel-level and semantic-level enhancements, respectively. FreqMoE utilizes the power of Mixture of LoRA Experts to adaptively handle diverse degradation types via FFT-based gating, while Orthogonal Gradient Projection ensures the semantic LoRA learns in a subspace orthogonal to the pixel-level representations, reducing subspace interference between fidelity and perceptual objectives.
  }
  \label{fig:architecture}
\end{figure*}

\subsection{Model Formulation}
\label{sec:formulation}
Multi-step DM-based SR methods~\cite{lin2024diffbir, wang2024exploiting, wu2024seesr} perform $T$-step denoising to transform Gaussian noise $z_T$ into the HR latent $z_H$, conditioned on $x_L$. This iterative process is computationally expensive and causes instability from random noise sampling. 
In this work, we adopt a one-step approach that starts directly from $z_L$ and leverages residual learning scheme: $z_H = z_L - \epsilon_\theta(z_L)$,
where $\epsilon_\theta$ is the SD UNet parameterized by $\theta$, and the noise schedule coefficients are absorbed into $\epsilon_\theta$ following~\cite{wu2024one, sun2025pixel}. This forces the network to learn the high-frequency difference between $z_L$ and $z_H$, accelerating training convergence. 

Following the dual-LoRA paradigm~\cite{sun2025pixel}, we introduce two sets of LoRA~\cite{hu2022lora} modules upon the frozen SD weights $\theta_{sd}$: a \textit{pixel-level} LoRA $\Delta\theta_{pix}$ trained with $\mathcal{L}_2$ loss for degradation removal, and a \textit{semantic-level} LoRA $\Delta\theta_{sem}$ trained with $\mathcal{L}_2$, $\mathcal{L}_{lpips}$~\cite{zhang2018unreasonable}, and $\mathcal{L}_{csd}$~\cite{yu2023text, sun2025pixel} losses for detail and semantic enhancement. The pixel-level LoRA is optimized first, then the semantic-level LoRA is trained while $\Delta\theta_{pix}$ remains frozen as follows:
\begin{align}
  z_H^{pix} &= z_L - \epsilon_{\theta_{pix}}(z_L), \quad \theta_{pix} = \{\theta_{sd},\, \Delta\theta_{pix}\}, \\
  z_H^{sem} &= z_L - \epsilon_{\theta_{full}}(z_L), \quad \theta_{full} = \{\theta_{sd},\, \Delta\theta_{pix},\, \Delta\theta_{sem}\}.
\end{align}
The decoded outputs $\hat{x}_H^{pix} = \mathcal{D}(z_H^{pix})$ and $\hat{x}_H^{sem} = \mathcal{D}(z_H^{sem})$ are used for loss computation in their respective phases.

Our overall framework is illustrated in Fig.~\ref{fig:architecture}, which extends dual-LoRA paradigm in two ways. First, we replace the single pixel-level LoRA $\Delta\theta_{pix}$ with a robust Frequency-guided Mixture of LoRA Experts (Sec.~\ref{sec:freqmoe}), enabling degradation-adaptive restoration rather than applying a static uniform approach. Second, beyond simply freezing $\Delta\theta_{pix}$ during semantic training, we additionally employ Orthogonal Gradient Projection (Sec.~\ref{sec:ogp}) to constrain semantic gradients to the null space of the pixel-level subspace, providing a mathematically stronger guarantee against fidelity degradation.

\subsection{Frequency-guided Mixture of LoRA Experts}
\label{sec:freqmoe}

\begin{figure}[t]
  \centering
  \begin{subfigure}[t]{0.49\linewidth}
    \centering
    \includegraphics[width=\linewidth]{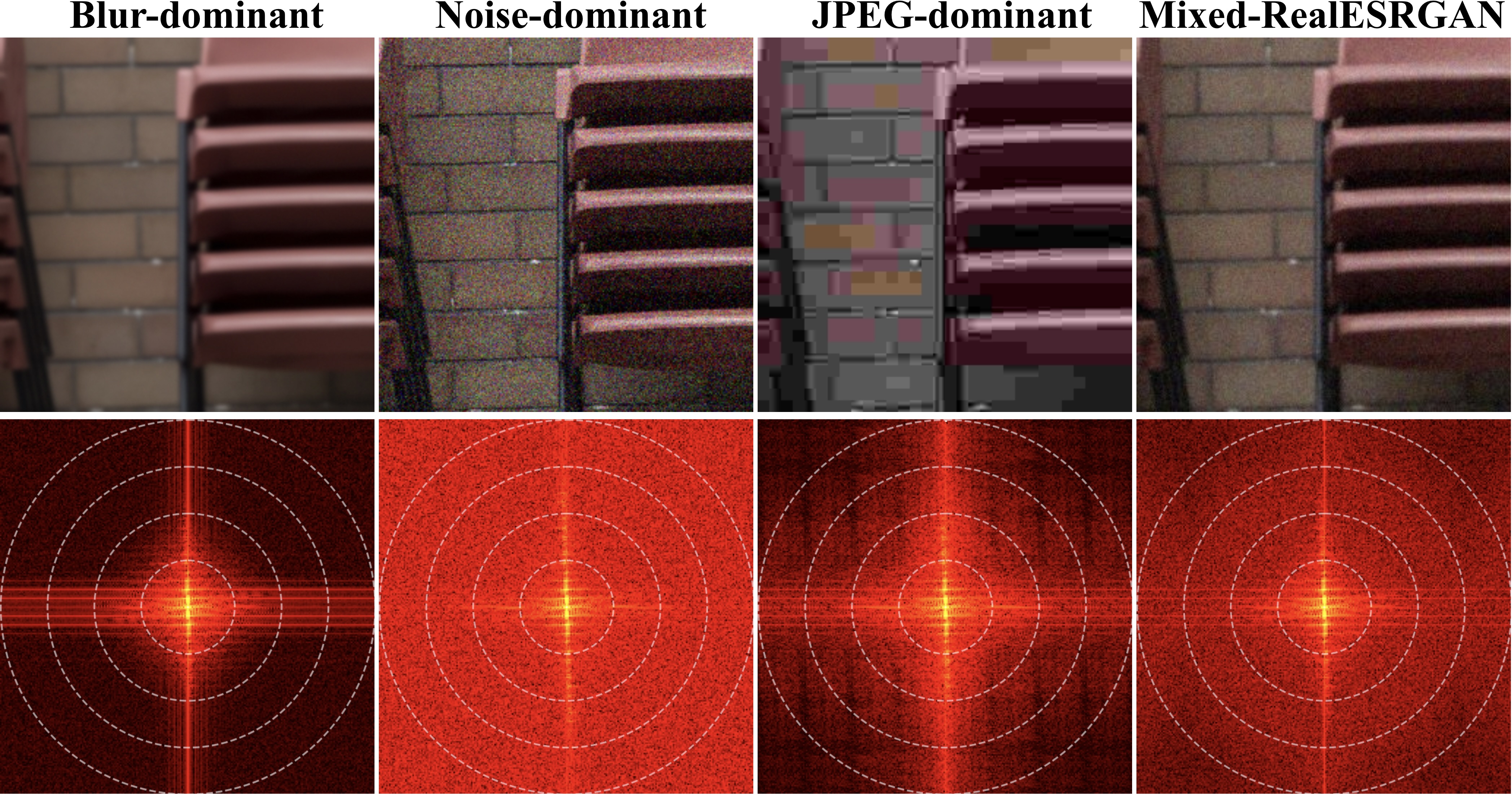}
    \caption{}
    \label{fig:freq_spectrum}
  \end{subfigure}
  \hfill
  \begin{subfigure}[t]{0.47\linewidth}
    \centering
    \includegraphics[width=\linewidth]{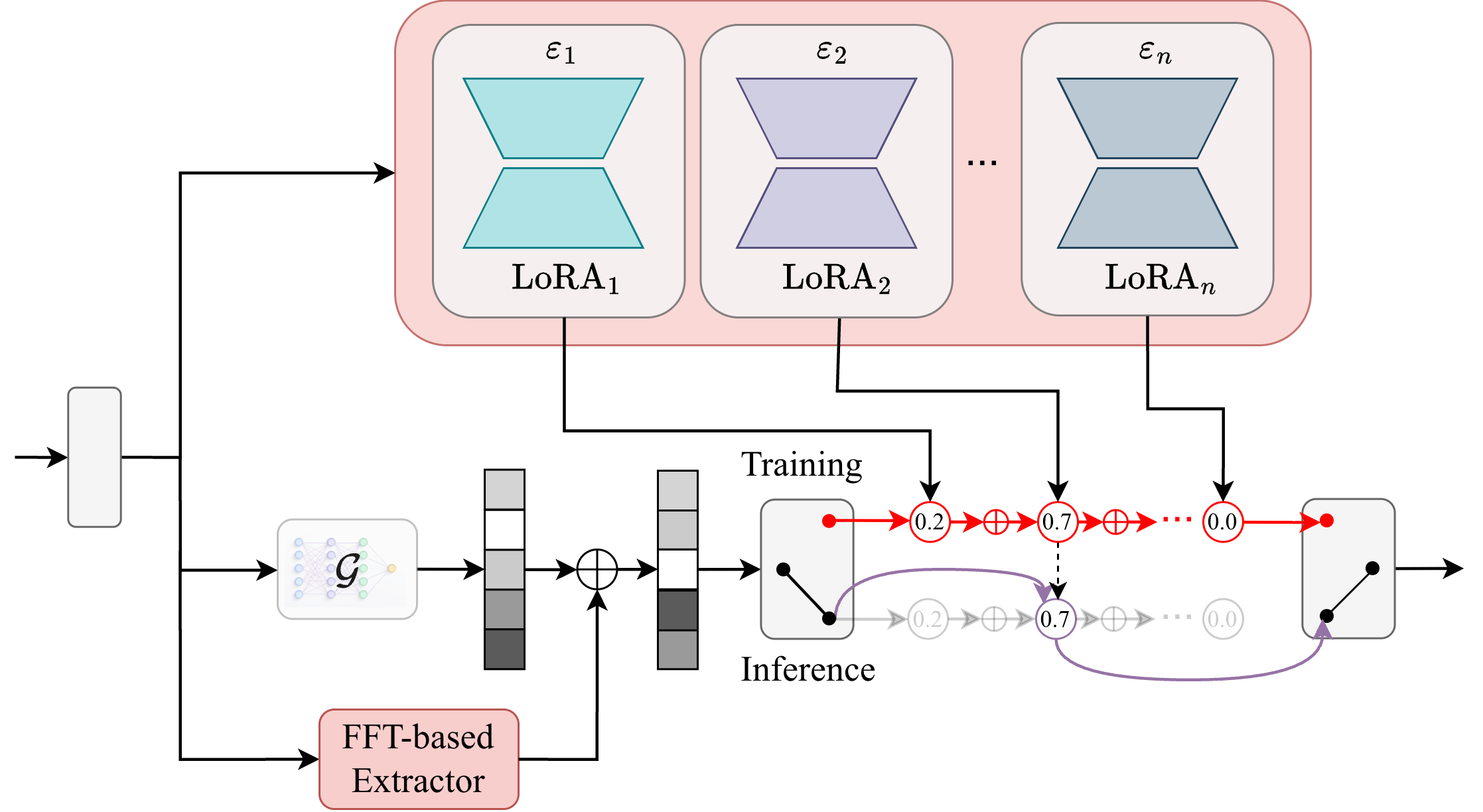}
    \caption{}
    \label{fig:freqmoe}
  \end{subfigure}
  \caption{(a) FFT magnitude spectra (log scale) of LR images under different degradations. Blur suppresses high-frequency energy (dim outer ring), noise raises it uniformly (bright outer ring), and JPEG introduces periodic block artifacts (grid pattern). These signatures enable our non-parametric extractor to reliably identify degradation types. (b) Structure of the FreqMoE module which routes inputs to specialized LoRA experts based on frequency-domain degradation features.}
  \label{fig:freqmoe_and_spectrum}
\end{figure}

Recent DM-based methods~\cite{wu2024one, sun2025pixel} employ a single LoRA adapter for pixel-level restoration, applying a uniform strategy regardless of degradation type. Since real-world images exhibit diverse, mixed degradations requiring distinct frequency-domain restoration behaviors (Fig.~\ref{fig:freq_spectrum}) \cite{wang2021real, wu2024seesr}, a single LoRA lacks the capacity to specialize across such heterogeneous patterns.
To address this challenge, we replace the monolithic pixel-level LoRA $\Delta\theta_{pix}$ with a Frequency-guided Mixture of LoRA Experts (FreqMoE; Fig.~\ref{fig:freqmoe}). Specifically, each target layer in the UNet now contains $N$ parallel LoRA expert pairs $\{(\mathbf{A}_i, \mathbf{B}_i)\}_{i=1}^{N}$, where $\mathbf{A}_i \in \mathbb{R}^{r \times d_{in}}$ and $\mathbf{B}_i \in \mathbb{R}^{d_{out} \times r}$ are the low-rank down- and up-projection matrices with rank $r$, respectively. A gating network $\mathcal{G}$ routes each input to the top-$k$ most suitable experts using both local spatial features from the layer input and global frequency-domain degradation features extracted from $x_L$. The FreqMoE output for a given layer is represented as follows:
\begin{equation}
  \Delta\mathbf{h}_{pix} = \sum_{i \in \text{Top-}k(\boldsymbol{\ell})} g_i(\mathbf{x}, \mathbf{f}) \cdot \mathbf{B}_i \mathbf{A}_i \mathbf{x},
  \label{eq:freqmoe}
\end{equation}
where $\mathbf{x}$ is the layer input, $\mathbf{f} \in \mathbb{R}^{d_f}$ is the degradation feature vector extracted from $x_L$, $\boldsymbol{\ell}$ are the gating logits, and $g_i(\mathbf{x}, \mathbf{f})$ is the routing weight for expert $i$ among the selected top-$k$ experts. This sparse routing reduces computational cost while enabling complementary experts to collaborate when beneficial.
Ablations on the number of experts, top-$k$ routing, and expert routing behavior are provided in Sec.~11 of the \textbf{supplementary materials}.

\subsubsection{Degradation feature extraction.}
\label{sec:deg_feat}

We design a lightweight, non-parametric feature extractor that captures degradation signatures from the LR input $x_L$ in the frequency domain. Real-world LR images are typically corrupted by a composition of multiple degradation types. Following the second-order degradation model of RealESRGAN~\cite{wang2021real}, which cascades blur, resize, noise, and JPEG compression in two sequential rounds, each corruption leaves a characteristic fingerprint in the frequency spectrum (Fig.~\ref{fig:freq_spectrum}): blur suppresses high-frequency content~\cite{kundur1996blind}, noise elevates energy uniformly across frequencies~\cite{buades2005review}, JPEG compression introduces periodic artifacts at $8{\times}8$ block boundaries~\cite{luo2010jpeg}, and resize operations produce aliasing patterns visible in the radial energy distribution~\cite{blu2004linear}.

Given an input image $x_L \in \mathbb{R}^{C \times H \times W}$, we first convert it to a grayscale image $x_g$ and compute its centered 2D FFT magnitude spectrum $\mathbf{M} = |\mathcal{F}(x_g)|$, where $\mathcal{F}$ is the shift-centered 2D FFT magnitude operator. The spectrum is divided into $K$ concentric radial bands $\{R_k\}_{k=1}^{K}$ from the center (low frequency) to the edges (high frequency), where each band spans the radial range $[(k{-}1)\rho/K,\; k\rho/K)$ and $\rho$ is the maximum radius. The normalized energy for each band is:
\begin{equation}
  \bar{M}_k = \frac{1}{|R_k|}\!\sum_{(u,v) \in R_k}\!\mathbf{M}(u,v), \quad
  e_k = \frac{\bar{M}_k}{\sum_{k'=1}^{K} \bar{M}_{k'}},
  \label{eq:freq_bands}
\end{equation}
where $|R_k|$ is the pixel count of band $R_k$. This area-normalized mean prevents larger outer bands from dominating, yielding a probability distribution $\mathbf{e} = [e_1, \ldots, e_K]$.
We further compute three scalar degradation indicators:
\begin{enumerate}

    \item \textit{Blur score}: $s_{blur} = e_1 / (e_K + \epsilon)$, measuring the ratio of low- to high-frequency energy. A high score indicates blur-typical low-frequency dominance.
    \item \textit{JPEG blocking score}: $s_{jpeg} = \bar{g}_{block} / (\bar{g}_{all} + \epsilon)$, where $\bar{g}_{block}$ is the mean absolute gradient at $8{\times}8$ block boundaries and $\bar{g}_{all}$ is the overall mean gradient. Elevated boundary gradients signal JPEG compression artifacts.
    \item \textit{Noise score}: $s_{noise} = \text{mean}(|\mathcal{L} * x_L|)$, where $\mathcal{L}$ is the $3{\times}3$ Laplacian kernel applied via depthwise convolution. A high Laplacian response indicates the presence of high-frequency noise.
    
\end{enumerate}
The final feature vector $\mathbf{f}\!=\![\mathbf{e};\, s_{blur};\, s_{jpeg};\, s_{noise}] \in \mathbb{R}^{d_f}$ ($d_f\!=\!K{+}3$) concatenates the band energies with the three scalar scores. While the three scalar scores target the dominant degradation types, the radial band energies $\mathbf{e}$ implicitly capture other corruptions such as resize-induced aliasing, which manifests as characteristic shifts in the radial energy distribution. Together, $\mathbf{f}$ provides a compact yet comprehensive degradation descriptor. This extractor is non-parametric and computed with gradients disabled, achieving negligible overhead.
The effect of the number of frequency bands $K$ is analyzed in Sec.~11 of the \textbf{supplementary materials}.

\subsubsection{Frequency-modulated gating network.}
\label{sec:freq_gate}

The gating network $\mathcal{G}$ combines spatial information from the layer activations with the global degradation features $\mathbf{f}$ to produce routing weights. Given the layer input $\mathbf{x} \in \mathbb{R}^{d_{in}}$ and degradation features $\mathbf{f} \in \mathbb{R}^{d_f}$, the gating logits are computed as:
\begin{equation}
  \boldsymbol{\ell} = \mathbf{W}_g \mathbf{x} + \phi(\mathbf{f}),
  \label{eq:gate}
\end{equation}
where $\mathbf{W}_g \in \mathbb{R}^{N \times d_{in}}$ is the base gate projection and $\phi: \mathbb{R}^{d_f} \to \mathbb{R}^{N}$ is a small MLP consisting of two linear layers with a ReLU activation: $\phi(\mathbf{f}) = \mathbf{W}_2 \, \text{ReLU}(\mathbf{W}_1 \mathbf{f})$, with $\mathbf{W}_1 \in \mathbb{R}^{2d_f \times d_f}$ and $\mathbf{W}_2 \in \mathbb{R}^{N \times 2d_f}$.

This additive design allows $\phi(\mathbf{f})$ to act as a degradation-de\-pendent bias that shifts the base routing logits. Since $\mathbf{f}$ is computed once per image and shared across all spatial positions and layers, the frequency bias provides a consistent global routing signal. Experts are not pre-assigned to specific degradation types; instead, they emerge through end-to-end training guided by the frequency features. For instance, experts may specialize in high-frequency recovery (activated for blurred inputs) or denoising (activated for noisy inputs), while mixed-degradation images activate multiple experts via top-$k$ routing.

For routing, we select the top-$k$ experts per token. The gating weights are obtained by applying softmax over the selected logits:
\begin{equation}
  g_i(\mathbf{x}, \mathbf{f}) =
  \begin{cases}
    \frac{\exp(\ell_i)}{\sum_{j \in \text{Top-}k} \exp(\ell_j)}, & \text{if } i \in \text{Top-}k(\boldsymbol{\ell}) \\
    0, & \text{otherwise}
  \end{cases}
  \label{eq:topk}
\end{equation}
$\text{Top-}k(\boldsymbol{\ell})$ returns the indices of $k$ largest logits. This sparse routing lowers computation over dense mixtures while still enabling complementary expert collaboration when useful.
For stable training, we initialize $\mathbf{W}_g$ with small random values ($\sigma{=}0.01$) and set the output layer of $\phi$ to zero, so the gating starts from near-uniform routing and gradually learns degradation-specific specialization.

To address expert collapse in MoE frameworks \cite{fedus2022switch, liu2024deepseek, do2025simsmoe}, we employ the Load Balancing Loss (LBL) \cite{fedus2022switch}, which encourages uniform expert utilization by linking the frequency of expert selection to the routing weights each expert receives.
Let \( f_i \) be the fraction of tokens directed to expert \( i \), \( P_i \) denote the average routing probability for expert \( i \), and \( N \) be the total number of experts. The LBL is defined as: $\mathcal{L}_{lbl} = \alpha \cdot N \cdot \sum_{i=1}^{N} f_i \cdot P_i.$

\subsection{Orthogonal Gradient Projection}
\label{sec:ogp}
\subsubsection{Problem formulation.}

A core challenge in decoupled real-world ISR is ensuring that distinct adapters capture truly independent features. We have identified a fundamental phenomenon in subspace interference: when a semantic-level adapter is trained on top of an existing fidelity-level module, the optimization process often collapses into the subspace already covered by the fidelity weights.

Formally, consider the joint inference state \(\theta_{full} = \{\theta_{sd}, \Delta\theta_{pix}, \Delta\theta_{sem}\}\). While \(\Delta\theta_{pix}\) is optimized for structural alignment, the subsequent optimization of \(\Delta\theta_{sem}\) lacks a mechanism to prevent it from re-learning structural features that are already encoded in \(\Delta\theta_{pix}\). If the update directions of the semantic LoRA are not constrained to be orthogonal to the fidelity subspace, the model suffers from representational redundancy. This redundancy wastes the model's limited rank capacity on repetitive structural information instead of capturing novel perceptual textures.
This prevents the model from reaching the optimal or near-optimal boundary of the perception-fidelity trade-off, irrespective of the parameter budget. To address this issue, we propose explicitly isolating these objectives by projecting semantic updates into the null space of the fidelity manifold.

\subsubsection{Subspace extraction via SVD.}

After the first phase, we extract the pixel-level subspace for each layer using Singular Value Decomposition (SVD) \cite{saha2021gradient}. For a layer with weight \(W\), the pixel LoRA introduces a change $\Delta W_{pix} = B_{pix} A_{pix} \in \mathbb{R}^{d_{out} \times d_{in}}$. In a MoE framework with \(N\) experts, we concatenate all expert weight changes to capture the union of their subspaces:
\[
\Delta W_{all} = [\Delta W_1 \mid \Delta W_2 \mid \dots \mid \Delta W_N] \in \mathbb{R}^{d_{out} \times (N \cdot d_{in})}
\]
We then perform SVD on \(\Delta W_{all} = U \Sigma V^T\) and retain the top \(\tilde{k}\) singular vectors that account for 95\% of the energy:

\[
\tilde{k} = \text{arg min}_{\tilde{k}} \left\{ \frac{\sum_{i=1}^{\tilde{k}} \sigma_i^2}{\sum_{j} \sigma_j^2} \geq 0.95 \right\}
\]

\noindent The truncated left singular matrix \(U_{\tilde{k}} \in \mathbb{R}^{d_{out} \times \tilde{k}}\) defines the essential pixel subspace preserved for Phase 2. This SVD is performed only once offline between training phases, not per iteration, and operates on relatively small matrices. On our hardware, the entire SVD extraction across all layers completes in under 5 seconds, introducing negligible overhead relative to the total training time.
Ablations on the SVD energy threshold are provided in Sec.~11 of the \textbf{supplementary materials}.

\subsubsection{Gradient projection during semantic LoRA training.}
During the second phase, we register backward hooks on all semantic LoRA $B$ weight matrices. When gradients are computed, the hook projects them into the null space of $U_{\tilde{k}}$:
\begin{equation}
  G_{ortho} = G - U_{\tilde{k}}(U_{\tilde{k}}^T G),
\end{equation}
\noindent where $G \in \mathbb{R}^{d_{out} \times r}$ is the original gradient of the LoRA $B$ weight. Since $U_{\tilde{k}}$ has orthonormal columns ($U_{\tilde{k}}^T U_{\tilde{k}} = I$), it follows that $U_{\tilde{k}}^T G_{ortho} = 0$, ensuring that the semantic gradient has zero component along the pixel-level subspace directions and removing pixel-subspace gradient interference. As a result, the semantic LoRA is guided to learn in a subspace independent of the pixel-level representations, enabling more effective perceptual enhancement. The projection is computationally efficient with negligible overhead.
The projection target choice is further justified and ablated in Sec.~11 of the \textbf{supplementary materials}.

\subsection{Training Strategy}

The loss design in both phases builds upon the combination of $\mathcal{L}_2$, $\mathcal{L}_{lpips}$, and $\mathcal{L}_{csd}$ losses, which has been shown effective for balancing pixel-level fidelity and semantic enhancement in diffusion-based SR~\cite{wu2024one, sun2025pixel}. We further use an intermediate SVD extraction step to enable orthogonal gradient projection.

\textbf{Phase 1 - FreqMoE Training.}
In the first phase, we train the FreqMoE module using the $\mathcal{L}_2$ loss to measure pixel-level fidelity between the predicted HQ image and ground truth, combined with the load balancing loss $\mathcal{L}_{lbl}$ (Sec.~\ref{sec:freqmoe}) to encourage balanced expert utilization. All FreqMoE parameters $\Delta\theta_{pix}$ are updated in this phase while the pre-trained SD parameters $(\theta_{sd})$ remain frozen.

\textbf{SVD-based subspace extraction.}
Between the two training phases, we extract the pixel-level subspace from the learned FreqMoE weights via SVD as described in Sec.~\ref{sec:ogp}. The resulting orthogonal bases $U_{\tilde{k}}$ are stored and used to construct the projection operators for Phase~2.

\textbf{Phase 2 - Semantic LoRA Training with OGP.}
In this phase, we train the semantic-level LoRA $\{\Delta\theta_{sem}\}$ while keeping both the pre-trained SD $\theta_{sd}$ and the pixel-level FreqMoE $(\Delta\theta_{pix})$ frozen. The training objective combines three complementary losses: $\mathcal{L}_{2}$ loss to maintain pixel-level consistency, $\mathcal{L}_{lpips}$ loss to align high-level features with a pre-trained VGG for perceptual quality, and $\mathcal{L}_{csd}$ loss to leverage semantic priors from the pre-trained SD for detail enhancement. Detailed loss configurations are provided in Sec.~6 of the \textbf{supplementary materials}.
Critically, during backpropagation, we employ the orthogonal gradient projection (Sec.~\ref{sec:ogp}) to all gradients flowing into the semantic LoRA $B$ matrices, ensuring the semantic LoRA learns in a subspace orthogonal to the pixel-level representations.

\subsubsection{Inference.}
In the default setting, both the FreqMoE and semantic LoRA are active, and the SR output is produced in a single forward pass via $z_H = z_L - \epsilon_{\theta_{full}}(z_L)$. 
% Inheriting from~\cite{sun2025pixel}, 
Our dual-LoRA design also supports an adjustable mode by introducing pixel-level and semantic-level guidance scales $\lambda_{pix}$ and $\lambda_{sem}$:
\begin{equation}
  \epsilon_\theta(z_L) = \lambda_{pix}\,\epsilon_{\theta_{pix}}(z_L) + \lambda_{sem}\bigl(\epsilon_{\theta_{full}}(z_L) - \epsilon_{\theta_{pix}}(z_L)\bigr),
  \label{eq:inference_adj}
\end{equation}
where $\epsilon_{\theta_{pix}}(z_L)$ is the output with only the FreqMoE active, and $\epsilon_{\theta_{full}}(z_L) - \epsilon_{\theta_{pix}}(z_L)$ isolates the semantic-level contribution. Setting $\lambda_{pix} {=} \lambda_{sem} {=} 1$ recovers the default single-pass mode. This adjustable mode requires two forward passes but enables controllable fidelity-perception trade-off without re-training.

\section{Experiments}
\label{sec:experiment}

\subsection{Experimental Settings}

\subsubsection{Training settings.} 
We train FreqOrtho-SR using the SD 2.1-base \cite{rombach2022high} for the $\times$4 SR task. A pixel-level FreqMoE and semantic LoRA modules are applied to the weights of all convolutional and MLP layers, both initialized using a Gaussian distribution with a rank of 4. The FreqMoE uses $N{=}4$ experts with top-$k{=}2$ routing, and the frequency band number is set to $K{=}4$ ($d_f{=}7$). 
Following recent methods \cite{wu2024one, sun2025pixel, yi2025fine, he2025mixture}, we use LSDIR \cite{li2023lsdir} and the first 10K images from FFHQ \cite{karras2019style} dataset as training data. We generate paired training data using RealESRGAN's degradation pipeline \cite{wang2021real}. The batch size and patch size are set to 16 and 512$\times$512. We use the AdamW optimizer \cite{kingma2014adam} with a learning rate of 5e-5. 
% and weight decay $10^{-2}$. 
We train Phase 1 and Phase 2 for 4K and 26K iterations, respectively.

\subsubsection{Compared methods.}
We compare FreqOrtho-SR with several recent one-step DM-based methods: AddSR \cite{tai2026addsr}, SinSR~\cite{wang2024sinsr}, OSEDiff~\cite{wu2024one}, MoR-DASR~\cite{he2025mixture}, PiSA-SR~\cite{sun2025pixel}, and TVT~\cite{yi2025fine}. Additional comparisons with GAN-based methods (BSRGAN~\cite{zhang2021designing}, RealESRGAN~\cite{wang2021real}, LDL~\cite{liang2022details}) and multi-step DM-based methods (DiffBIR \cite{lin2024diffbir}, PASD \cite{yang2024pixel}, SeeSR \cite{wu2024seesr}) are provided in Sec.~7 and Sec.~8 of the \textbf{supplementary materials}, respectively. All comparative results are obtained from officially released codes.

\subsubsection{Test datasets and metrics.}
Following prior work \cite{sun2025pixel, he2025mixture}, we test on both synthetic and real-world data. Synthetic test samples are obtained by applying RealESRGAN's degradation pipeline \cite{wang2021real} to DIV2K dataset \cite{agustsson2017ntire}. Real-world data are center-cropped from the RealSR \cite{cai2019toward} and DRealSR \cite{wei2020component} datasets.
% , with LR images sized at 128$\times$128 pixels and HR images sized at 512$\times$512 pixels.
To measure fidelity, we use PSNR and SSIM \cite{wang2004image}, computed on the Y channel in YCbCr color space. We also evaluate perceptual quality using LPIPS \cite{zhang2018unreasonable} and DISTS \cite{ding2020image}, both computed in RGB space. FID \cite{heusel2017gans} is used to evaluate the distribution similarity between GT and SR images. For image quality assessment without reference GT, we use NIQE \cite{mittal2012making}, CLIPIQA \cite{wang2023exploring}, MUSIQ \cite{ke2021musiq}, and MANIQA \cite{yang2022maniqa}.

%%%%%%%%%%%%%%%%%%%%%%%%%%%%%%%%%%%%%%%%%%%%%%%%%%%%%%% MAIN TABLE %%%%%%%%%%%%%%%%%%%%%%%%%%%%%%%%%%%%%%%%%%%%%%%%%%%%%%%
\begin{table}[t]
\centering
\setlength{\extrarowheight}{0pt}
\addtolength{\extrarowheight}{\aboverulesep}
\addtolength{\extrarowheight}{\belowrulesep}
\setlength{\aboverulesep}{0pt}
\setlength{\belowrulesep}{0pt}
\caption{Quantitative comparison among the state-of-the-art one-step DM-based SR methods on synthetic and real-world test datasets. The best and the second-best results are highlighted in \textcolor{best}{red} and \textcolor{second}{blue}, respectively. ``-'' indicates metrics not reported in the original paper due to unavailable code.}
\label{tab:main_table}
\setlength{\tabcolsep}{0.65mm}
% \scalebox{0.72}{
{\fontsize{6.pt}{6pt}\selectfont
\begin{tabular}{clccccccccc}
\toprule
Datasets & Methods      & PSNR$\uparrow$ & SSIM$\uparrow$ & LPIPS$\downarrow$ & DISTS$\downarrow$ & FID$\downarrow$ & NIQE$\downarrow$ & CLIPIQA$\uparrow$ & MUSIQ$\uparrow$ & MANIQA$\uparrow$ \\
\midrule
%====================================== RealSR ======================================
\multirow{7}{*}{RealSR}
    & AddSR     & 23.12     & 0.6550     & 0.3090   & 0.2703     & 154.14     & 6.66     & 0.5520    & 67.14 &    0.4880 \\
    & SinSR     & \textcolor{red}{26.30}   & 0.7354                    & 0.3212                     & 0.2346                     & 137.05                     & 6.31                      & 0.6204                     & 60.41                     & 0.5389                      \\
    & OSEDiff   & 25.15                    & 0.7341                    & 0.2921                     & 0.2128                     & 123.50                     & 5.65                      & 0.6693                     & 69.09                     & 0.6339                      \\
    & MoR-DASR  & 25.32                    & 0.7280                    & 0.2910                     & -                          & -                          & -                         & \textcolor{red}{0.6910}    & 69.78                     & 0.512                       \\
    & PiSA-SR   & 25.50                    & 0.7417                    & 0.2672                     & \textcolor{blue}{0.2044}   & 124.09                     & \textcolor{blue}{5.50}                      & 0.6702                     & \textcolor{red}{70.15}    & \textcolor{blue}{0.6560}    \\
    & TVT       & 25.81                    & \textcolor{red}{0.7596}   & \textcolor{blue}{0.2587}   & 0.2061                     & \textcolor{blue}{109.93}   & 5.92                      & \textcolor{blue}{0.6882}   & \textcolor{blue}{69.89}   & 0.6228                      \\
    \rowcolor{gray!10}
    \rowcolor{gray!10}
    & FreqOrtho-SR   & \textcolor{blue}{26.27}  & \textcolor{blue}{0.7481}  & \textcolor{red}{0.2537}    & \textcolor{red}{0.1951}    & \textcolor{red}{108.91}    & \textcolor{red}{5.32}                      & 0.6630                     & 69.39                     & \textcolor{red}{0.6586}     \\
\cline{1-1}\cmidrule{2-11}
%====================================== DRealSR ======================================
\multirow{7}{*}{DRealSR}
    & AddSR     & 26.71     & 0.7380        & 0.3210    & 0.2631     & 164.79     & 7.72     & 0.5930    & 62.13     & 0.4580 \\
    & SinSR     & \textcolor{blue}{28.41}  & 0.7495                    & 0.3741                     & 0.2488                     & 177.05                     & 7.02                      & 0.6367                     & 55.34                     & 0.4898                      \\
    & OSEDiff   & 27.92                    & 0.7835                    & 0.2968                     & \textcolor{blue}{0.2165}                     & 135.29                     & 6.49                      & 0.6963                     & 64.65                     & 0.5899                      \\
    & MoR-DASR  & 28.37                    & 0.7760                    & 0.3070                     & -                          & -                          & -                         & \textcolor{blue}{0.7170}   & \textcolor{blue}{65.94}   & 0.5090                      \\
    & PiSA-SR   & 28.31                    & 0.7804                    & 0.2960                     & 0.2169                     & \textcolor{red}{130.61}    & \textcolor{red}{6.20}                      & 0.6970                     & \textcolor{red}{66.11}    & \textcolor{blue}{0.6156}   \\
    & TVT       & 28.27                    & \textcolor{blue}{0.7899}  & \textcolor{blue}{0.2900}   & 0.2205                     & 134.19                     & 7.02                      & \textcolor{red}{0.7220}    & 65.56                     & 0.5783                      \\
    \rowcolor{gray!10}
    \rowcolor{gray!10}
    & FreqOrtho-SR   & \textcolor{red}{29.08}   & \textcolor{red}{0.7911}   & \textcolor{red}{0.2799}    & \textcolor{red}{0.2110}    & \textcolor{blue}{130.85}   & \textcolor{blue}{6.29}                      & 0.6879                     & 65.89                     & \textcolor{red}{0.6182}     \\
    \cline{1-1}\cmidrule{2-11}

    %====================================== DIV2K ======================================
\multirow{7}{*}{DIV2K}
    & AddSR     & 23.26     & 0.5900    & 0.3620    & 0.2344     & 35.03     & 5.82    & 0.5730     & 63.39     & 0.4050 \\
    & SinSR     & \textcolor{red}{24.43}   & 0.6012                    & 0.3262                     & 0.2066                     & 35.45                      & 6.02                      & 0.6499                     & 62.80                     & 0.5395                      \\
    & OSEDiff   & 23.72                    & 0.6108                    & 0.2941                     & 0.1976                     & 26.32                      & 4.71                      & 0.6683                     & 67.97                     & 0.6148                      \\
    & MoR-DASR  & 24.01                    & 0.6060                    & 0.2890                     & -                          & -                          & -                         & 0.6810                     & 68.09                     & 0.4750                      \\
    & PiSA-SR   & 23.87                    & 0.6058                    & 0.2823                     & 0.1934                     & 25.07                      & \textcolor{red}{4.55}    & \textcolor{blue}{0.6927}   & \textcolor{red}{69.68}    & \textcolor{red}{0.6400}    \\
    & TVT       & 24.23                    & \textcolor{red}{0.6292}   & \textcolor{blue}{0.2773}   & \textcolor{red}{0.1860}    & \textcolor{blue}{24.78}    & 5.60                      & \textcolor{red}{0.6986}    & \textcolor{blue}{68.67}                     & 0.6037                      \\
    \rowcolor{gray!10}
    \rowcolor{gray!10}
    & FreqOrtho-SR   & \textcolor{blue}{24.35}  & \textcolor{blue}{0.6189}  & \textcolor{red}{0.2748}    & \textcolor{blue}{0.1889}   & \textcolor{red}{23.54}     & \textcolor{blue}{4.63}                      & 0.6702                     & \textcolor{blue}{68.67}   & \textcolor{blue}{0.6355}    \\ \bottomrule
\end{tabular}}
\end{table}
%%%%%%%%%%%%%%%%%%%%%%%%%%%%%%%%%%%%%%%%%%%%%%%%%%%%%%% MAIN TABLE %%%%%%%%%%%%%%%%%%%%%%%%%%%%%%%%%%%%%%%%%%%%%%%%%%%%%%%

\subsection{Comparisons with State-of-the-Art Methods}

\subsubsection{Quantitative results.}
Table~\ref{tab:main_table} compares the performance of our FreqOrtho-SR with SOTA one-step DM-based SR methods across both synthetic and real-world test datasets. From this comparison, we can make the following observations:

\begin{itemize}
\setlength{\itemsep}{0pt}
\setlength{\topsep}{2pt}
    \item First, our method demonstrates clear advantages over competing methods in full-reference fidelity metrics, such as PSNR and SSIM, as well as in perceptual quality metrics including LPIPS and DISTS. Notably, these advantages are especially evident on the two real-world datasets, DRealSR and RealSR. 

    \item Second, FreqOrtho-SR consistently performs well on non-reference metrics like NIQE and MANIQA, which are widely recognized for evaluating real-world ISR. Although PiSA-SR surpasses us on one non-reference metric (MUSIQ), our performance on this metric remains comparable. This is satisfactory given that we excel on most of the metrics listed in Table~\ref{tab:main_table}.
\end{itemize}

\noindent Notably, FreqOrtho-SR manages to achieve competitive results on both fidelity and perceptual quality metrics, which is a significant challenge in real-world ISR. This is achieved through our proposed components: Frequency-guided Mixture of LoRA Experts and orthogonal gradient projection, in a two-phase training setup that mimics a sequential task setup.
% This represents a significant challenge for methods targeting real-world ISR.
Generally, there is no single method that consistently improves performance at all metrics; results vary substantially across metrics. In contrast, FreqOrtho-SR achieves the best or second-best in several cases, indicating substantial improvements in the real-world ISR field.

\subsubsection{Qualitative results.}
\begin{figure}[t]
\centering
\includegraphics[width=0.93\linewidth]{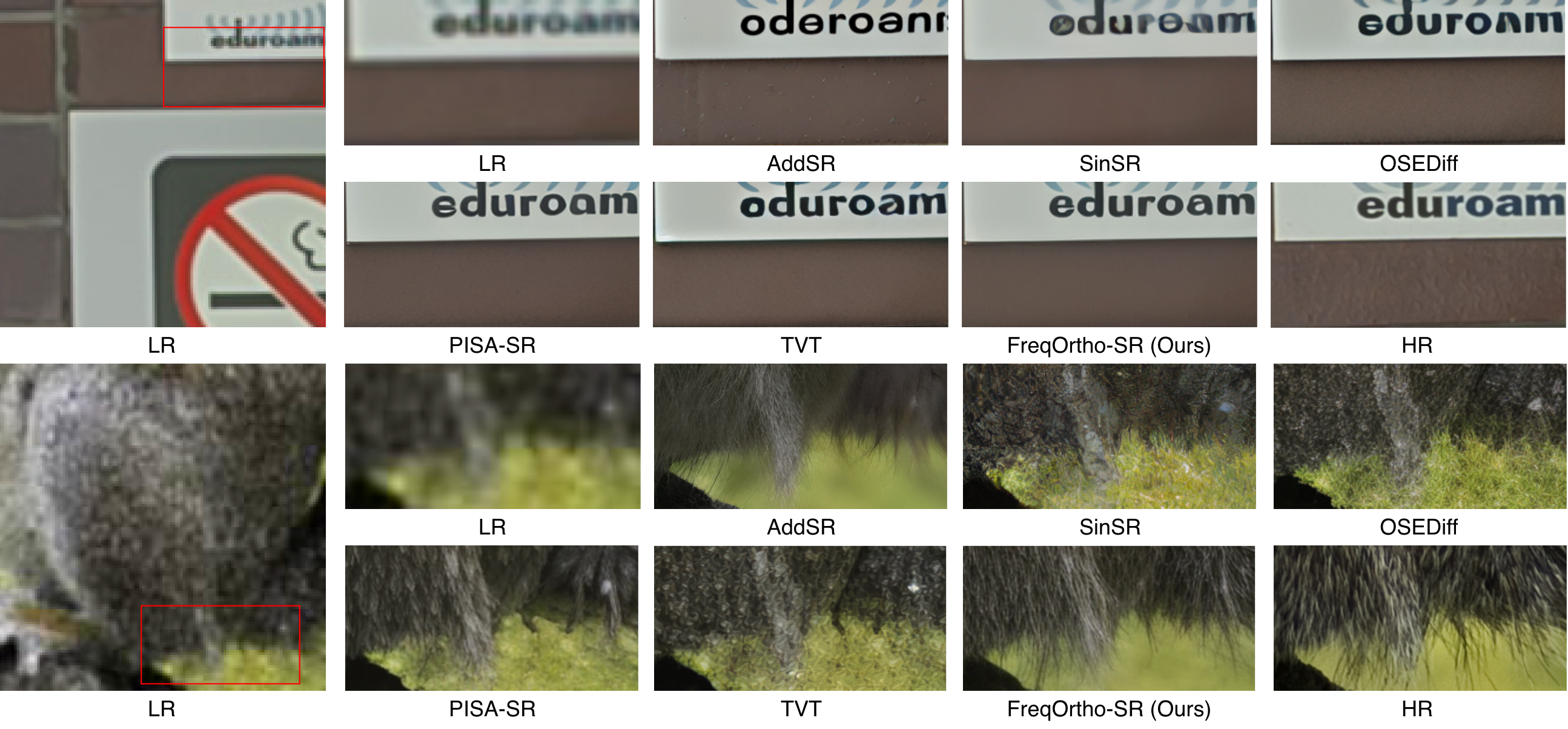}
\caption{Visual comparisons of one-step DM-based SR methods. Zoom in for best view.}
\label{fig:qualitative}
\end{figure}

Fig.~\ref{fig:qualitative} presents visual comparisons of FreqOrtho-SR with SOTA one-step DM-based SR methods.
Add\-SR~\cite{tai2026addsr} and Sin\-SR~\cite{wang2024sinsr} produce over-smoothed textures, failing to recover fine-grained details.
OSEDiff~\cite{wu2024one} generates more consistent outputs but with limited semantic details.
While PiSA-SR~\cite{sun2025pixel} and TVT~\cite{yi2025fine} can restore richer textures, they occasionally introduce visual artifacts or structural distortions.
In contrast, FreqOrtho-SR reconstructs more accurate structures (\eg, the sharp text in the first example) and produces more natural, realistic details (\eg, the fine textures in the second example), benefiting from frequency-guided expert learning that effectively disentangles pixel-level and semantic-level enhancements.
As described in Sec.~\ref{sec:methodology}, our dual-LoRA design naturally supports an adjustable inference mode (Eq.~\eqref{eq:inference_adj}), enabling controllable fidelity-perception trade-off via pixel-semantic guidance scales. Detailed adjustable SR experiments are provided in Sec.~9 of the \textbf{supplementary materials}.

\subsubsection{Complexity analysis.}
\begin{wraptable}[12]{r}{0.45\textwidth}
\vspace{-10pt}
\centering
\caption{Complexity and performance of one-step DM-based methods. Time on $128{\times}128$ inputs (A100). LPIPS/FID on RealSR. $\dagger$: no code.}
\label{tab:complexity}
\setlength{\tabcolsep}{2pt}
\scalebox{0.68}{
\begin{tabular}{lcccc}
\toprule
Method & Time (s) & Params (B) & LPIPS$\downarrow$ & FID$\downarrow$ \\
\midrule
SinSR            & 0.13 & 0.18 & 0.3121 & 137.05 \\
OSEDiff          & 0.12 & 1.77 & 0.2921 & 123.50 \\
MoR-DASR$\dagger$& -    & -    & 0.2910 & -      \\
PiSA-SR          & 0.09 & 1.30 & 0.2672 & 124.09 \\
TVT              & 0.28 & 1.72 & 0.2587 & 109.93 \\
\textbf{Ours}    & 0.30 & 1.30 & \textcolor{best}{0.2537} & \textcolor{best}{108.91} \\
\bottomrule
\end{tabular}}
\vspace{-10pt}
\end{wraptable}
Table~\ref{tab:complexity} reports run time, model size, and representative metrics of recent SOTA methods on RealSR dataset.
FreqOrtho-SR has same parameter count as PiSA-SR (1.30\,B); lower than OSEDiff (1.77\,B) and TVT (1.72\,B). 
For FLOPs/peak memory, PiSA-SR, TVT, and FreqOrtho-SR require 2.23T/3.00\,GB, 2.97T/7.31\,GB, and 2.25T/3.03\,GB, respectively.
Thus, our activated compute and memory remain close to PiSA-SR, while being much lower than TVT in memory.

The additional inference time stems from 
FreqMoE routing: unlike a single LoRA that
can merge into 
the base weights for free, MoE structure requires computing top-$k$ experts per token along with the lightweight FFT-based gating, yielding a higher run time.
Despite this overhead, FreqOrtho-SR has comparable inference time compared to the current SOTA model TVT \cite{yi2025fine} while achieving the best LPIPS and FID among all compared methods, yielding a favorable quality-efficiency trade-off.
We note that MoE inference overhead can potentially be reduced via expert pruning or structured sparsity that we leave for future work.

\subsubsection{OOD real-world evaluation.}
To further evaluate robustness to unknown real-world degradations, we test on RealLR200 \cite{wu2024seesr}, a no-reference real LR benchmark without ground-truth HR images.
As shown in Table~\ref{tab:ood_reallr200}, FreqOrtho-SR achieves the best no-reference scores among the strongest one-step baselines, supporting its generalization beyond the RealESRGAN degradation pipeline.

\begin{table}[t]
\centering
\caption{OOD evaluation on RealLR200 \cite{wu2024seesr} (real-world, no GT). The image shows an in-the-wild LR input (left) and our output (right).}
\label{tab:ood_reallr200}
\setlength{\tabcolsep}{1.5pt}
\begin{minipage}[c]{0.44\linewidth}
  \centering
  \includegraphics[width=\linewidth]{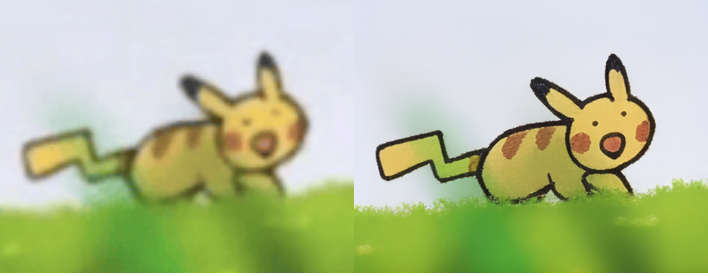}
\end{minipage}
\hfill
\begin{minipage}[c]{0.52\linewidth}
  \centering
  \scriptsize
  \begin{tabular}{lcccc}
  \toprule
  Method & NIQE$\downarrow$ & CLIPIQA$\uparrow$ & MUSIQ$\uparrow$ & MANIQA$\uparrow$ \\
  \midrule
  PiSA-SR & 4.00 & 0.6869 & 70.5058 & 0.6271 \\
  TVT     & 4.73 & 0.6909 & 70.5275 & 0.5954 \\
  \textbf{Ours} & \textcolor{best}{3.90} & \textcolor{best}{0.6912} & \textcolor{best}{70.6040} & \textcolor{best}{0.6321} \\
  \bottomrule
  \end{tabular}
\end{minipage}
\end{table}

\subsection{Ablation Study}

We conduct ablation studies on the RealSR dataset to validate all proposed components of FreqOrtho-SR (Table~\ref{tab:ablation}) by progressively removing each component. For a fair comparison, we run all experiments on the same standard settings.
Additional branch-design ablations, including applying FreqMoE to both the pixel and semantic branches, are reported in Sec.~11 of the \textbf{supplementary materials}.

FreqMoE with spatial-only gating (i.e., without frequency features) in the second row already improves fidelity (PSNR $+0.87$\,dB, SSIM $+0.015$), showing the benefit of multi-expert capacity. Adding frequency-guided routing (first row) further improves LPIPS and FID, confirming that FFT-based degradation signatures provide a more discriminative routing signal than spatial features alone. OGP (third row) also independently reduces FID, even without multi-expert routing. The complete FreqOrtho-SR (last row), which combines FreqMoE and OGP, achieves the best DISTS, FID, NIQE, and MANIQA scores among all variants, with a slight PSNR trade-off as OGP constrains the semantic LoRA to an independent subspace. The significant FID improvement confirms better output distribution, validating that OGP effectively mitigates subspace-level redundancy.

\begin{center}
\begin{minipage}{\linewidth}
\centering
\captionof{table}{Ablations on the RealSR dataset. We validate the effectiveness of each proposed component. The best results are highlighted in \textcolor{best}{red}.}
\label{tab:ablation}
\setlength{\tabcolsep}{2.5mm}
\scalebox{0.70}{
\begin{tabular}{ccccccccccc}
\toprule
MoE & Freq & OGP & PSNR$\uparrow$ & SSIM$\uparrow$ & LPIPS$\downarrow$ & DISTS$\downarrow$ & FID$\downarrow$ & NIQE$\downarrow$ & MANIQA$\uparrow$ \\
\midrule
$\surd$  & $\surd$  & $\times$ & 26.3097 & 0.7520 & \textcolor{best}{0.2517} & 0.1968 & 111.45 & 5.4812 & 0.6510 \\
$\surd$  & $\times$ & $\times$ & \textcolor{best}{26.3718} & \textcolor{best}{0.7571} & 0.2584 & 0.2017 & 118.35 & 5.678  & 0.6560 \\
$\times$ & $\times$ & $\surd$  & 26.1992 & 0.7432 & 0.2651 & 0.2021 & 113.39 & 5.4390 & 0.6557 \\
\rowcolor{gray!10}
$\surd$  & $\surd$  & $\surd$  & 26.27   & 0.7481 & 0.2537 & \textcolor{best}{0.1951} & \textcolor{best}{108.91} & \textcolor{best}{5.3200} & \textcolor{best}{0.6586} \\
\bottomrule
\end{tabular}}
\end{minipage}
\end{center}

\enlargethispage{4\baselineskip}
\subsubsection{Subspace orthogonality analysis.}
\begin{wrapfigure}[9]{r}{0.30\textwidth}
\vspace{-6pt}
\centering
\includegraphics[width=\linewidth]{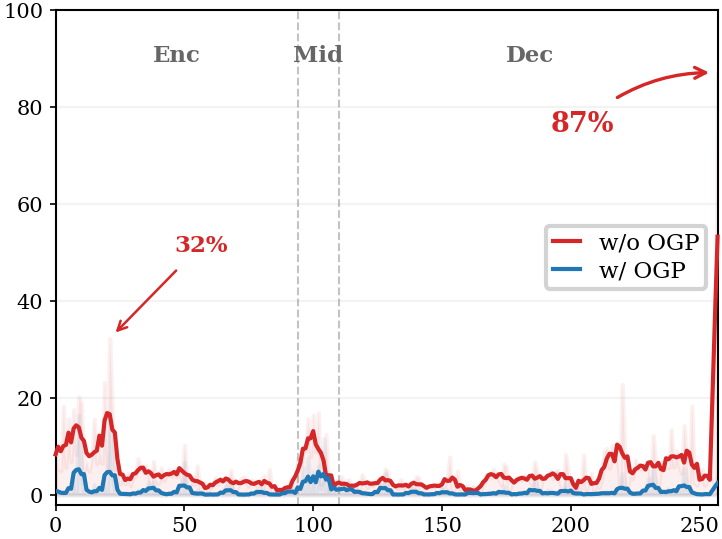}
\caption{Per-layer projection energy ratio onto the pixel-level subspace.}
\label{fig:subspace_overlap}
\vspace{-10pt}
\end{wrapfigure}
To directly verify that OGP enforces subspace independence and explain the perceptual gains when adding OGP to FreqMoE in Table~\ref{tab:ablation} (row 1 vs.\ last row), we measure the per-layer projection energy ratio $\frac{\|U_{\tilde{k}}^{(l)\top} \Delta W_{sem}^{(l)}\|_F^2}{\|\Delta W_{sem}^{(l)}\|_F^2}$, which quantifies how much of the semantic LoRA weights lie within the pixel-level subspace (lower is better). As shown in Fig.~\ref{fig:subspace_overlap}, the overlap without OGP is highly non-uniform: the output layer reaches 87\% and encoder attention layers up to 32\%, meaning nearly all semantic capacity at those layers redundantly duplicates pixel-level features. OGP suppresses the overlap to near-zero across all 258 layers, confirming that the semantic LoRA learns in a truly independent subspace, thereby improving perceptual metrics DISTS, FID, NIQE, and MANIQA simultaneously.

\section{Conclusion}
We propose FreqOrtho-SR, a one-step diffusion framework that introduces two novel principles for real-world image super-resolution.
FreqMoE introduces fre\-quency-do\-main degradation signatures as a principled routing signal for expert specialization, yielding stable and interpretable gating without additional learned feature extractors.
OGP bridges continual learning with multi-objective SR optimization, providing a provable orthogonality guarantee that moves beyond simple parameter freezing.
Together, these contributions demonstrate a strong fidelity-perception balance across RealSR, DRealSR, and DIV2K, with competitive or best performance on key metrics among one-step diffusion methods.
\section*{Acknowledgements}
\vspace{-0.4em}
{\small\sloppy
This work was supported by the National Research Foundation of Korea (NRF) grant funded by the Korea government (MSIT) (RS-2025-00573160); the Technology Innovation Program (RS-2025-02222776, Development and Demonstration of AI and Lightweight Technology-Based Automated E-Waste Sorting and Retrieval System) funded by the Ministry of Trade, Industry and Resources (MOTIR, Korea); the IITP (Institute of Information \& Communications Technology Planning \& Evaluation)-ITRC (Information Technology Research Center) grant funded by the Korea government (Ministry of Science and ICT) (IITP-2026-RS-2023-00259703); and the ``Advanced GPU Utilization Support Program'' funded by the Government of the Republic of Korea (Ministry of Science and ICT).

This work was also supported by Hyundai Motor Chung Mong-Koo Global Scholarship to Dinh Phu Tran (co-first author, equal contribution).

\par}


\bibliographystyle{splncs04}
\bibliography{main}

\end{document}

% ---------------------------------------------------------------
\title{FreqOrtho-SR: Frequency-Guided Orthogonal Expert Learning for Real-World Image Super-Resolution \\
{\large \normalfont Supplementary Material}} 

\titlerunning{FreqOrtho-SR}

\author{Minh Son Hoang\thanks{Equal contribution.} \and
Dinh Phu Tran\textsuperscript{$\star$} \and
Quyen Nguyen Duc \and
Dam Hoang Phuong \and
Daeyoung Kim\thanks{Corresponding author.}}

\authorrunning{M.~S.~Hoang, D.~P.~Tran et al.}

\institute{School of Computing, KAIST, Republic of Korea\\
\email{\{sonhm2910,phutx2000,nguyenducquyen2311,hoangphuong1211,kimd\}@kaist.ac.kr}}

\maketitle

% Continue numbering from main paper
\setcounter{section}{5}
\setcounter{figure}{5}
\setcounter{table}{3}
\setcounter{equation}{8}

\textit{This supplementary material provides additional implementation details, theoretical analysis, extended comparisons, and visualizations for the main paper ``FreqOrtho-SR: Frequency-Guided Orthogonal Expert Learning for Real-World Image Super-Resolution.''}

\noindent We organize the supplementary as follows:

\begin{itemize}
    \item \textbf{Sec.~\ref{sec:impl_details}}: Implementation details including training configurations, hardware setup, and hyperparameters.
    \item \textbf{Sec.~\ref{sec:gan},~\ref{sec:multistep}}: Extended comparisons with GAN-based and multi-step DM-based methods.
    \item \textbf{Sec.~\ref{sec:adjustable_sr}}: Experiments on adjustable SR with pixel-semantic guidance scales.
    \item \textbf{Sec.~\ref{sec:proof_ogp}}: Distinction between our orthogonal gradient projection and continual-learning orthogonal projection methods.
    \item \textbf{Sec.~\ref{sec:more_ablation}}: Additional ablation studies on SVD energy threshold, number of experts and top-$k$, FreqMoE expert routing, number of frequency bands, OGP projection target, and effect of applying FreqMoE to both branches.
    \item \textbf{Sec.~\ref{sec:more_qualitative}}: More qualitative results.
    \item \textbf{Sec.~\ref{sec:limitations}}: Limitations and future works.
\end{itemize}

\section{Implementation Details}
\label{sec:impl_details}

In this section, we provide additional training details that complement Sec.~4.1 of the main paper, which covers the experimental settings. For fair comparison, we use the training settings (e.g., training data, degradation pipeline, patch size, and evaluation protocol) from recent one-step diffusion-based SR works~\cite{sun2025pixel,wu2024one,yu2024scaling}. While the two-phase training strategy, loss functions, and architectural choices are discussed in the main paper, we now detail the hardware setup, optimizer configuration, and remaining hyperparameters.

As described in Sec.~3.4 of the main paper, training proceeds in two phases. In \textbf{Phase~1 (Pixel LoRA Training)}, the FreqMoE module is trained with:
\begin{equation}
  \mathcal{L}_{\text{phase1}} = \lambda_{l2}\,\mathcal{L}_2 + \lambda_{lbl}\,\mathcal{L}_{lbl}.
\end{equation}
After Phase~1, we extract the pixel-level subspace via SVD (Sec.~3.3 of the main paper). In \textbf{Phase~2 (Semantic LoRA Training with OGP)}, the semantic LoRA is trained under orthogonal gradient projection with:
\begin{equation}
  \mathcal{L}_{\text{phase2}} = \lambda_{l2}\,\mathcal{L}_2 + \lambda_{lpips}\,\mathcal{L}_{lpips} + \lambda_{csd}\,\mathcal{L}_{csd}.
\end{equation}

All training is conducted on 8 NVIDIA A100 (40\,GB) GPUs with FP16 mixed precision using the Accelerate framework. We use the AdamW optimizer with $\beta_1{=}0.9$, $\beta_2{=}0.999$, $\epsilon{=}10^{-8}$, and weight decay $10^{-2}$. The learning rate is set to $5 \times 10^{-5}$ with a constant schedule after 500 warmup steps, shared across both phases. In Phase~2, the CFG scale is set to 1.0 with timestep ratios sampled from $[0.02, 0.98]$.
Table~\ref{tab:hyperparams} provides a complete summary of all hyperparameters.

\begin{table}[t]
\centering
\caption{Summary of key hyperparameters for FreqOrtho-SR.}
\label{tab:hyperparams}
\setlength{\tabcolsep}{4pt}
\scalebox{0.78}{
\begin{tabular}{lll}
\toprule
Category & Parameter & Value \\
\midrule
\multirow{6}{*}{Training}
  & Optimizer & AdamW ($\beta_1{=}0.9$, $\beta_2{=}0.999$, $\epsilon{=}10^{-8}$) \\
  & Learning rate & $5 \times 10^{-5}$ (constant, 500 warmup steps) \\
  & Weight decay & $10^{-2}$ \\
  & Batch size & 16 (2 per device $\times$ 8 GPUs) \\
  & Training iterations & 4K (Phase 1) + 26K (Phase 2) \\
  & Mixed precision & FP16 \\
\midrule
\multirow{5}{*}{FreqMoE}
  & Number of experts ($N$) & 4 \\
  & Top-$k$ routing & 2 \\
  & LoRA rank (pixel) & 4 \\
  & Number of frequency bands ($K$) & 4 \\
  & Frequency feature dimension ($d_f$) & 7 ($K{+}3$) \\
\midrule
\multirow{2}{*}{Semantic LoRA}
  & LoRA rank (semantic) & 4 \\
  & SVD energy threshold ($\tau$) & 0.95 \\
\midrule
\multirow{4}{*}{Loss weights}
  & $\lambda_{l2}$ (both phases) & 1.0 \\
  & $\lambda_{lpips}$ (Phase 2 only) & 2.0 \\
  & $\lambda_{csd}$ (Phase 2 only) & 1.0 \\
  & $\lambda_{lbl}$ (Phase 1 only) & 0.01 \\
\bottomrule
\end{tabular}}
\end{table}

\section{Comparison with GAN-based Methods}
\label{sec:gan}

We compare FreqOrtho-SR with three GAN-based SR methods: RealESRGAN~\cite{wang2021real}, BSRGAN~\cite{zhang2021designing}, and LDL~\cite{liang2022details} in Table~\ref{tab:gan}.
FreqOrtho-SR achieves the best no-reference metrics (NIQE, CLIPIQA, MUSIQ, MANIQA) and the best LPIPS across all three datasets, while maintaining competitive fidelity (PSNR, SSIM) and DISTS.

\begin{table}[h!]
\centering
\setlength{\extrarowheight}{0pt}
\addtolength{\extrarowheight}{\aboverulesep}
\addtolength{\extrarowheight}{\belowrulesep}
\setlength{\aboverulesep}{0pt}
\setlength{\belowrulesep}{0pt}
\caption{Quantitative comparison with GAN-based SR methods on synthetic and real-world test datasets. The best and the second-best results are highlighted in \textcolor{best}{red} and \textcolor{second}{blue}, respectively.}
\label{tab:gan}
\setlength{\tabcolsep}{1mm}
\scalebox{0.72}{
\begin{tabular}{clccccccccc}
\toprule
Datasets & Methods      & PSNR$\uparrow$ & SSIM$\uparrow$ & LPIPS$\downarrow$ & DISTS$\downarrow$ & FID$\downarrow$ & NIQE$\downarrow$ & CLIPIQA$\uparrow$ & MUSIQ$\uparrow$ & MANIQA$\uparrow$ \\
\midrule
%====================================== RealSR ======================================
\multirow{4}{*}{RealSR}
    & RealESRGAN   & 25.69 & \textcolor{second}{0.7616} & 0.2727 & \textcolor{second}{0.2063} & \textcolor{second}{135.18} & 5.83 & 0.4449 & 60.18 & \textcolor{second}{0.5487} \\
    & BSRGAN       & \textcolor{best}{26.39} & \textcolor{best}{0.7654} & \textcolor{second}{0.2670} & 0.2121 & 141.28 & \textcolor{second}{5.66} & \textcolor{second}{0.5001} & \textcolor{second}{63.21} & 0.5399 \\
    & LDL          & 25.28 & 0.7567 & 0.2766 & 0.2121 & 142.71 & 6.00 & 0.4477 & 60.82 & 0.5485 \\
    \cmidrule{2-11}
    \rowcolor{gray!10}
    & Ours & \textcolor{second}{26.27} & 0.7481 & \textcolor{best}{0.2537} & \textcolor{best}{0.1951} & \textcolor{best}{108.91} & \textcolor{best}{5.32} & \textcolor{best}{0.6630} & \textcolor{best}{69.39} & \textcolor{best}{0.6586} \\
\cline{1-1}\cmidrule{2-11}
%====================================== DRealSR ======================================
\multirow{4}{*}{DRealSR}
    & RealESRGAN   & 28.64 & \textcolor{second}{0.8053} & 0.2847 & \textcolor{best}{0.2089} & \textcolor{second}{147.62} & 6.69 & 0.4422 & 54.18 & 0.4907 \\
    & BSRGAN       & \textcolor{second}{28.75} & 0.8031 & 0.2883 & 0.2142 & 155.63 & \textcolor{second}{6.52} & \textcolor{second}{0.4915} & \textcolor{second}{57.14} & 0.4878 \\
    & LDL          & 28.21 & \textcolor{best}{0.8126} & \textcolor{second}{0.2815} & 0.2132 & 155.53 & 7.13 & 0.4310 & 53.85 & \textcolor{second}{0.4914} \\
    \cmidrule{2-11}
    \rowcolor{gray!10}
    & Ours & \textcolor{best}{29.08} & 0.7911 & \textcolor{best}{0.2799} & \textcolor{second}{0.2110} & \textcolor{best}{130.85} & \textcolor{best}{6.29} & \textcolor{best}{0.6879} & \textcolor{best}{65.89} & \textcolor{best}{0.6182} \\
\cline{1-1}\cmidrule{2-11}
%====================================== DIV2K ======================================
\multirow{4}{*}{DIV2K}
    & RealESRGAN   & 24.29 & \textcolor{best}{0.6371} & \textcolor{second}{0.3112} & \textcolor{second}{0.2141} & \textcolor{second}{37.64} & \textcolor{second}{4.68} & \textcolor{second}{0.5277} & 61.06 & \textcolor{second}{0.5501} \\
    & BSRGAN       & \textcolor{best}{24.58} & 0.6269 & 0.3351 & 0.2275 & 44.23 & 4.75 & 0.5071 & \textcolor{second}{61.20} & 0.5247 \\
    & LDL          & 23.83 & \textcolor{second}{0.6344} & 0.3256 & 0.2227 & 42.29 & 4.85 & 0.5180 & 60.04 & 0.5350 \\
    \cmidrule{2-11}
    \rowcolor{gray!10}
    & Ours & \textcolor{second}{24.35} & 0.6189 & \textcolor{best}{0.2748} & \textcolor{best}{0.1889} & \textcolor{best}{23.54} & \textcolor{best}{4.63} & \textcolor{best}{0.6702} & \textcolor{best}{68.67} & \textcolor{best}{0.6355} \\
\bottomrule
\end{tabular}}
\end{table}

\section{Comparison with Multi-step DM-based Methods}
\label{sec:multistep}

We compare FreqOrtho-SR with multi-step DM-based SR methods: DiffBIR~\cite{lin2024diffbir} (50 steps), PASD~\cite{yang2024pixel} (20 steps), and SeeSR~\cite{wu2024seesr} (50 steps) in Table~\ref{tab:multistep}.
FreqOrtho-SR, using just one diffusion step, achieves the highest fidelity (PSNR, SSIM) and reference-based perceptual metrics (LPIPS, DISTS, FID) across all three datasets.
On the two real-world datasets, RealSR and DRealSR, which consist of real photographic images, FreqOrtho-SR achieves the highest scores across most no-reference quality metrics.
On the synthetic DIV2K dataset, which consists entirely of artificially generated images for super-resolution benchmarking, multi-step methods such as PASD and SeeSR achieve slightly better no-reference scores (CLIPIQA, MUSIQ, MANIQA). This is expected, as their iterative refinement across many steps progressively synthesizes richer texture details, yielding higher no-reference quality scores regardless of fidelity to the ground truth~\cite{wu2024one}, as evidenced by their inferior full-reference metrics.
Moreover, FreqOrtho-SR also allows a controllable fidelity-perception trade-off through its guidance scales (see Sec.~\ref{sec:adjustable_sr}), enabling users to boost perceptual quality as needed, unlike less flexible multi-step methods.
Taken together, these results indicate that single-step FreqOrtho-SR offers a favorable trade-off between restoration quality and computational efficiency, particularly for real-world scenarios where inference cost is a practical concern alongside quantitative performance.

\begin{table}[h!]
\centering
\setlength{\extrarowheight}{0pt}
\addtolength{\extrarowheight}{\aboverulesep}
\addtolength{\extrarowheight}{\belowrulesep}
\setlength{\aboverulesep}{0pt}
\setlength{\belowrulesep}{0pt}
\caption{Quantitative comparison with multi-step DM-based SR methods on synthetic and real-world test datasets. The best and the second-best results are highlighted in \textcolor{best}{red} and \textcolor{second}{blue}, respectively.}
\label{tab:multistep}
\setlength{\tabcolsep}{1mm}
\scalebox{0.72}{
\begin{tabular}{clccccccccc}
\toprule
Datasets & Methods      & PSNR$\uparrow$ & SSIM$\uparrow$ & LPIPS$\downarrow$ & DISTS$\downarrow$ & FID$\downarrow$ & NIQE$\downarrow$ & CLIPIQA$\uparrow$ & MUSIQ$\uparrow$ & MANIQA$\uparrow$ \\
\midrule
%====================================== RealSR ======================================
\multirow{4}{*}{RealSR}
    & DiffBIR-S50  & 24.88 & 0.6673 & 0.3567 & 0.2290 & 124.56 & 5.63 & 0.6412 & 64.66 & 0.6231 \\
    & PASD-S20     & 25.22 & 0.6809 & 0.3392 & 0.2259 & \textcolor{second}{123.08} & \textcolor{best}{5.18} & 0.6502 & 68.74 & \textcolor{second}{0.6461} \\
    & SeeSR-S50    & \textcolor{second}{25.33} & \textcolor{second}{0.7273} & \textcolor{second}{0.2985} & \textcolor{second}{0.2213} & 125.66 & 5.38 & \textcolor{second}{0.6594} & \textcolor{second}{69.37} & 0.6439 \\
    \cmidrule{2-11}
    \rowcolor{gray!10}
    & Ours & \textcolor{best}{26.27} & \textcolor{best}{0.7481} & \textcolor{best}{0.2537} & \textcolor{best}{0.1951} & \textcolor{best}{108.91} & \textcolor{second}{5.32} & \textcolor{best}{0.6630} & \textcolor{best}{69.39} & \textcolor{best}{0.6586} \\
\cline{1-1}\cmidrule{2-11}
%====================================== DRealSR ======================================
\multirow{4}{*}{DRealSR}
    & DiffBIR-S50  & 26.84 & 0.6660 & 0.4446 & 0.2706 & 167.38 & \textcolor{second}{6.02} & 0.6292 & 60.68 & 0.5902 \\
    & PASD-S20     & 27.48 & 0.7051 & 0.3854 & 0.2535 & 157.36 & \textcolor{best}{5.57} & \textcolor{second}{0.6714} & 64.55 & \textcolor{second}{0.6130} \\
    & SeeSR-S50    & \textcolor{second}{28.26} & \textcolor{second}{0.7698} & \textcolor{second}{0.3197} & \textcolor{second}{0.2306} & \textcolor{second}{149.86} & 6.52 & 0.6672 & \textcolor{second}{64.84} & 0.6026 \\
    \cmidrule{2-11}
    \rowcolor{gray!10}
    & Ours & \textcolor{best}{29.08} & \textcolor{best}{0.7911} & \textcolor{best}{0.2799} & \textcolor{best}{0.2110} & \textcolor{best}{130.85} & 6.29 & \textcolor{best}{0.6879} & \textcolor{best}{65.89} & \textcolor{best}{0.6182} \\
\cline{1-1}\cmidrule{2-11}
%====================================== DIV2K ======================================
\multirow{4}{*}{DIV2K}
    & DiffBIR-S50  & 23.67 & 0.5653 & 0.3541 & 0.2129 & 30.93 & 4.71 & 0.6652 & 65.66 & 0.6204 \\
    & PASD-S20     & 23.14 & 0.5489 & 0.3607 & 0.2219 & 29.32 & \textcolor{best}{4.40} & \textcolor{second}{0.6711} & \textcolor{best}{68.83} & \textcolor{best}{0.6484} \\
    & SeeSR-S50    & \textcolor{second}{23.71} & \textcolor{second}{0.6045} & \textcolor{second}{0.3207} & \textcolor{second}{0.1967} & \textcolor{second}{25.83} & 4.82 & \textcolor{best}{0.6857} & 68.49 & 0.6239 \\
    \cmidrule{2-11}
    \rowcolor{gray!10}
    & Ours & \textcolor{best}{24.35} & \textcolor{best}{0.6189} & \textcolor{best}{0.2748} & \textcolor{best}{0.1889} & \textcolor{best}{23.54} & \textcolor{second}{4.63} & 0.6702 & \textcolor{second}{68.67} & \textcolor{second}{0.6355} \\
\bottomrule
\end{tabular}}
\end{table}

\section{Experiments on Adjustable SR}
\label{sec:adjustable_sr}

We validate the adjustable inference capability of FreqOrtho-SR by fixing one guidance scale ($\lambda_{pix}$ or $\lambda_{sem}$) at 1.0 and varying the other on the RealSR test dataset (Table~\ref{tab:adjustable_sr}).
PSNR and SSIM measure pixel-level fidelity, LPIPS and DISTS assess perceptual similarity, and FID evaluates distributional distance to the GT, while NIQE, CLIPIQA, MUSIQ, and MANIQA are no-reference image quality metrics.

\noindent\textbf{Effect of $\lambda_{pix}$.}
Increasing $\lambda_{pix}$ progressively removes degradations and enhances edges, leading to a continuous improvement in no-reference metrics.
However, the reference-based metrics exhibit a rise-and-fall pattern: PSNR peaks at $\lambda_{pix}=0.5$ (26.95\,dB), and LPIPS is minimized at $\lambda_{pix}=0.8$ (0.2490), indicating the best perceptual similarity to the GT.
Further increasing $\lambda_{pix}$ causes over-smoothing, degrading both PSNR and LPIPS.

\noindent\textbf{Effect of $\lambda_{sem}$.}
Increasing $\lambda_{sem}$ synthesizes richer semantic details, yielding a higher upper bound on no-reference metrics than pixel-level adjustments (MUSIQ reaches 71.38 and MANIQA reaches 0.6970 at $\lambda_{sem}=1.5$).
Meanwhile, PSNR generally decreases, and LPIPS first improves and peaks at $\lambda_{sem}=0.8$ (0.2427) before deteriorating, as excessive semantic enhancement introduces content deviations from the GT.
These results confirm that FreqOrtho-SR offers effective and controllable fidelity-perception trade-off to accommodate diverse user preferences.

\begin{center}
\begin{minipage}{\linewidth}
\centering
\captionof{table}{Results of FreqOrtho-SR with different pixel-semantic guidance scales on the RealSR test dataset.}
\label{tab:adjustable_sr}
\setlength{\tabcolsep}{3pt}
\scalebox{0.8}{
\begin{tabular}{cc ccccccccc}
\toprule
\cellcolor{purple!20}$\lambda_{pix}$ & \cellcolor{green!20}$\lambda_{sem}$ & PSNR$\uparrow$ & SSIM$\uparrow$ & LPIPS$\downarrow$ & DISTS$\downarrow$ & FID$\downarrow$ & NIQE$\downarrow$ & CLIPIQA$\uparrow$ & MUSIQ$\uparrow$ & MANIQA$\uparrow$ \\
\midrule
% --- Fix lambda_sem = 1.0, vary lambda_pix ---
\cellcolor{purple!12}0.0 & 1.0 & 26.1976 & 0.7114 & 0.3110 & 0.2106 & 137.07 & 4.5874 & 0.3717 & 46.8565 & 0.4440 \\
\cellcolor{purple!12}0.2 & 1.0 & 26.6115 & 0.7310 & 0.2831 & 0.1955 & 123.32 & 4.6710 & 0.4297 & 52.9328 & 0.4841 \\
\cellcolor{purple!12}0.5 & 1.0 & 26.9534 & 0.7507 & 0.2533 & 0.1862 & 107.22 & 5.0093 & 0.5214 & 61.2985 & 0.5557 \\
\cellcolor{purple!12}0.8 & 1.0 & 26.7719 & 0.7553 & 0.2490 & 0.1908 & 105.64 & 5.2900 & 0.6015 & 67.1483 & 0.6229 \\
\cellcolor{purple!12}1.0 & 1.0 & 26.27 & 0.7481 & 0.2537 & 0.1951 & 108.91 & 5.3200 & 0.6630 & 69.39 & 0.6586 \\
\cellcolor{purple!12}1.2 & 1.0 & 25.7816 & 0.7413 & 0.2620 & 0.1986 & 117.53 & 5.5887 & 0.6869 & 69.7382 & 0.6758 \\
\cellcolor{purple!12}1.5 & 1.0 & 24.8135 & 0.7238 & 0.2715 & 0.2011 & 125.90 & 6.2119 & 0.6735 & 69.4345 & 0.6805 \\
\midrule
% --- Fix lambda_pix = 1.0, vary lambda_sem ---
1.0 & \cellcolor{green!12}0.0 & 27.7503 & 0.7972 & 0.2925 & 0.2344 & 148.59 & 9.0197 & 0.3119 & 51.0414 & 0.4089 \\
1.0 & \cellcolor{green!12}0.2 & 27.7616 & 0.7971 & 0.2754 & 0.2252 & 143.51 & 8.8588 & 0.3585 & 53.7260 & 0.4448 \\
1.0 & \cellcolor{green!12}0.5 & 27.5683 & 0.7913 & 0.2491 & 0.2054 & 128.33 & 7.8136 & 0.4342 & 58.3332 & 0.5344 \\
1.0 & \cellcolor{green!12}0.8 & 27.0207 & 0.7740 & 0.2427 & 0.1931 & 114.67 & 6.1587 & 0.5757 & 65.3324 & 0.6159 \\
1.0 & \cellcolor{green!12}1.0 & 26.27 & 0.7481 & 0.2537 & 0.1951 & 108.91 & 5.3200 & 0.6630 & 69.39 & 0.6586 \\
1.0 & \cellcolor{green!12}1.2 & 25.4375 & 0.7151 & 0.2797 & 0.2043 & 112.02 & 5.1726 & 0.7009 & 70.7966 & 0.6824 \\
1.0 & \cellcolor{green!12}1.5 & 23.8350 & 0.6501 & 0.3239 & 0.2212 & 120.19 & 5.6872 & 0.6879 & 71.3780 & 0.6970 \\
\bottomrule
\end{tabular}}
\end{minipage}
\end{center}

\begin{center}
\begin{minipage}{\linewidth}
\centering
\includegraphics[width=0.98\linewidth]{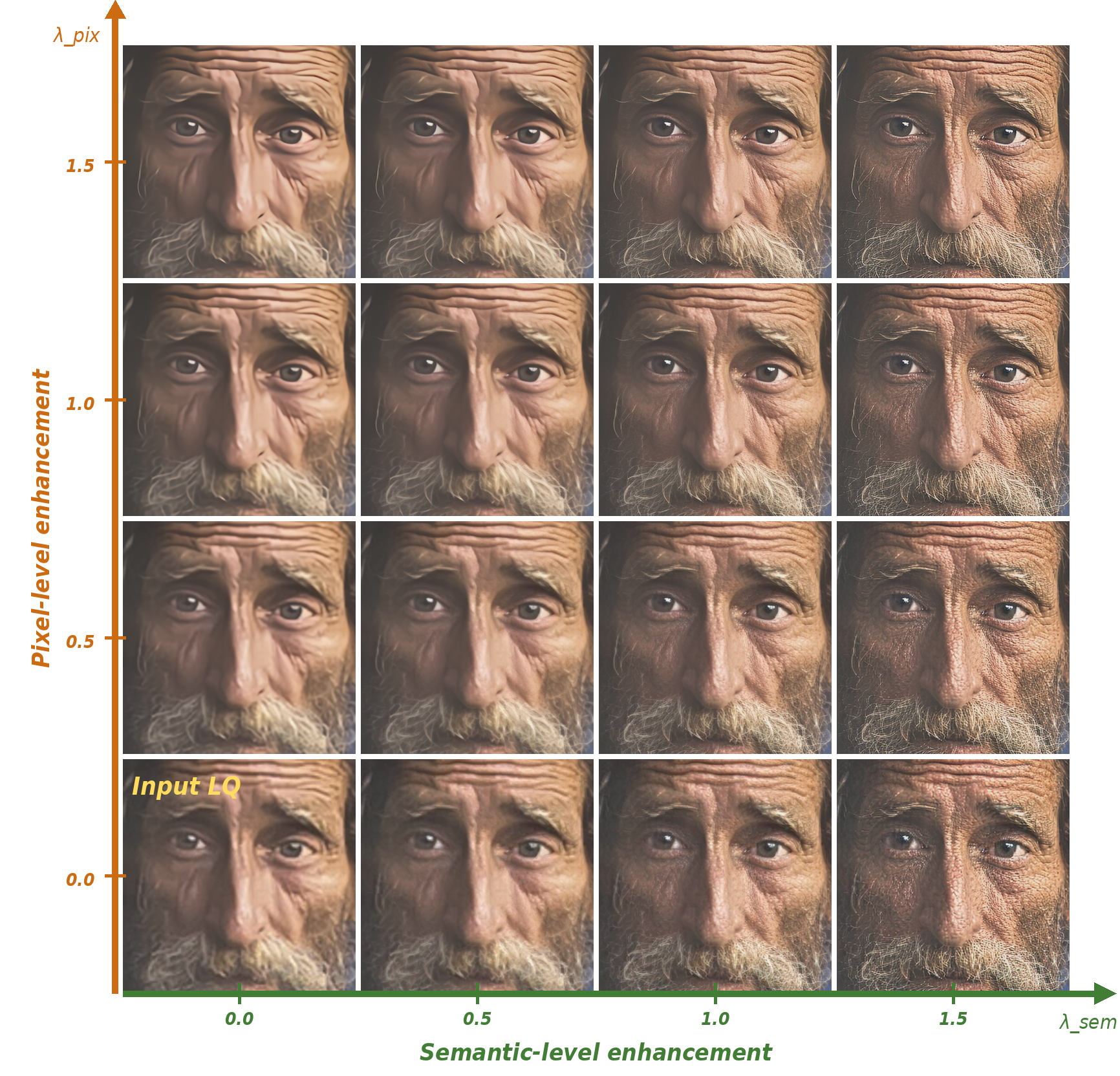}
\captionof{figure}{Qualitative adjustable SR results with different pixel-semantic guidance scales. The vertical axis varies the pixel-level guidance scale $\lambda_{pix}$, while the horizontal axis varies the semantic-level guidance scale $\lambda_{sem}$.}
\label{fig:adjustable_sr_grid}
\end{minipage}
\end{center}

\noindent\textbf{Qualitative analysis.}
Fig.~\ref{fig:adjustable_sr_grid} further visualizes the effect of the two guidance scales on an in-the-wild low-quality input.
Increasing $\lambda_{pix}$ mainly strengthens low-level restoration: blur is progressively reduced and facial structures become more stable, but overly large pixel-level guidance tends to suppress fine stochastic textures.
In contrast, increasing $\lambda_{sem}$ enriches semantic details such as wrinkles, skin texture, and beard strands, producing sharper and more realistic results.
However, excessive semantic-level guidance may also amplify synthesized details beyond the input evidence.
This qualitative behavior is consistent with Table~\ref{tab:adjustable_sr}: $\lambda_{pix}$ primarily controls fidelity-oriented restoration, whereas $\lambda_{sem}$ provides a stronger handle for perceptual enhancement, enabling users to select the desired fidelity-perception trade-off at inference time.

\section{Distinction from Continual-Learning Orthogonal Projection}
\label{sec:proof_ogp}

Orthogonal gradient projection is a well-established technique in continual learning~\cite{zeng2019continual,farajtabar2020orthogonal,saha2021gradient,wang2021training}, where it is used to prevent catastrophic forgetting: the subspace of previous tasks is identified from \emph{input activations} or \emph{accumulated gradient directions}, and new-task gradients are projected into the null space of that subspace so that previously learned knowledge is preserved.

While the projection formula,
\begin{equation}\label{eq:ogp_projection}
  G_{ortho} = G - U_{\tilde{k}}(U_{\tilde{k}}^T G),
\end{equation}
is a standard orthogonal complement operation, our approach differs from the continual-learning setting in three key aspects:

\noindent\textbf{(i) Subspace from weight deltas of MoE experts, not from activations or gradients.}
Continual-learning methods typically construct the protected subspace from input activations~\cite{zeng2019continual} or gradient outer products~\cite{farajtabar2020orthogonal} accumulated over training. These proxies characterize \emph{what the network has seen}, not \emph{what the network has learned to do}. In contrast, we first concatenate the weight deltas across all $N$ FreqMoE experts and then perform SVD to extract the pixel-level subspace:
\begin{equation}
  \Delta W_{all} = [\Delta W_1 \mid \cdots \mid \Delta W_N] \in \mathbb{R}^{d_{out} \times (N \cdot d_{in})}, \quad
  \Delta W_{all} = U \Sigma V^T,
\end{equation}
where the top $\tilde{k}$ left singular vectors $U_{\tilde{k}} \in \mathbb{R}^{d_{out} \times \tilde{k}}$ capture the \emph{output directions that the pixel-level module has learned to modify}, providing a direct characterization of the fidelity subspace and serving as the projection basis $U_{\tilde{k}}$ in Eq.~\eqref{eq:ogp_projection}. Moreover, concatenation naturally accounts for the union of all expert subspaces, which is specific to our MoE architecture and has no counterpart in standard continual-learning formulations.

\noindent\textbf{(ii) LoRA-aware projection: $B$-only sufficiency.}
Because our adapters use the LoRA parameterization $\Delta W = BA$, orthogonality of the full weight change,
\begin{equation}
  U_{\tilde{k}}^T \Delta W_{sem} = U_{\tilde{k}}^T (B_{sem} A_{sem}) = (U_{\tilde{k}}^T B_{sem})\, A_{sem} = \mathbf{0},
\end{equation}
is guaranteed by projecting \emph{only the $B$ matrix gradients}, regardless of $A$ (see Sec.~\ref{sec:abl_ogp_target} for the formal argument and ablation). This structural insight, arising from the low-rank factorization, is absent in continual-learning methods that operate on full-rank weight matrices.

\noindent\textbf{(iii) Objective: specialization, not forgetting prevention.}
Taken together, the above technical choices reflect a fundamentally different objective. In continual learning, projection protects previously learned tasks from being overwritten. In our setting, the pixel-level FreqMoE and the semantic LoRA are \emph{co-existing branches that operate simultaneously} at inference. The projection enforces that these two branches capture complementary information rather than collapsing into the same subspace, directly addressing the representational redundancy problem identified in Sec.~3.3 of the main paper.

\section{More Ablation Studies}
\label{sec:more_ablation}

\subsection{Effect of SVD Energy Threshold (Projection Ratio)}
\label{sec:abl_projection_ratio}

The SVD energy threshold $\tau$ controls how many singular vectors define the pixel-level subspace $U_{\tilde{k}}$, where $\tilde{k}$ denotes the average number of retained singular vectors across all LoRA layers. A higher $\tau$ retains more singular vectors (larger $\tilde{k}$), yielding a more complete description of the pixel-level subspace and thus a stricter orthogonal constraint on semantic LoRA updates. Results on the RealSR dataset are shown in Table~\ref{tab:abl_projection_ratio}.

\begin{table}[t]
\centering
\caption{Ablation study on SVD energy threshold (projection ratio) of the OGP module on the RealSR dataset. The threshold $\tau$ controls the fraction of singular value energy retained for the pixel-level subspace. The best results are highlighted in \textcolor{best}{red}.}
\label{tab:abl_projection_ratio}
\setlength{\tabcolsep}{2pt}
\scalebox{0.78}{
\begin{tabular}{cccccccccccc}
\toprule
$\tau$ & $\tilde{k}$ (avg.) & PSNR$\uparrow$ & SSIM$\uparrow$ & LPIPS$\downarrow$ & DISTS$\downarrow$ & FID$\downarrow$ & NIQE$\downarrow$ & CLIPIQA$\uparrow$ & MUSIQ$\uparrow$ & MANIQA$\uparrow$ \\
\midrule
0.80 & 5.78 & 26.34 & 0.7517 & \textcolor{best}{0.2526} & \textcolor{best}{0.1950} & 112.93 & 5.49 & 0.6556 & 68.72 & 0.6529 \\
0.85 & 6.68 & 26.25 & 0.7515 & 0.2540 & 0.1971 & 113.17 & 5.49 & 0.6594 & 68.86 & 0.6548 \\
0.90 & 7.79 & 26.35 & 0.7508 & 0.2620 & 0.2003 & 115.52 & 5.70 & 0.6621 & 68.76 & 0.6507 \\
\rowcolor{gray!10}
0.95 & 9.20 & 26.27 & 0.7481 & 0.2537 & 0.1951 & \textcolor{best}{108.91} & \textcolor{best}{5.32} & \textcolor{best}{0.6630} & \textcolor{best}{69.39} & \textcolor{best}{0.6586} \\
0.99 & 11.48 & \textcolor{best}{26.55} & \textcolor{best}{0.7579} & 0.2545 & 0.1975 & 114.22 & 5.57 & 0.6496 & 68.62 & 0.6500 \\
\bottomrule
\end{tabular}}
\end{table}

The results reveal a clear and systematic trade-off controlled by $\tau$. Lower thresholds ($\tau{=}0.80$, $\tilde{k}{\approx}5.78$) impose a looser orthogonal constraint, allowing the semantic LoRA to partially overlap with the pixel-level subspace; this preserves slightly better fidelity metrics (best LPIPS and DISTS) but limits perceptual improvement. As $\tau$ increases, the constraint becomes stricter and the semantic LoRA is forced to learn in a truly independent subspace, progressively improving perceptual quality. At $\tau{=}0.95$ ($\tilde{k}{\approx}9.20$), the model achieves the best FID (108.91), NIQE (5.32), and all no-reference scores (CLIPIQA, MUSIQ, MANIQA), while the fidelity degradation remains marginal compared to the best. Interestingly, pushing the threshold further to $\tau{=}0.99$ ($\tilde{k}{\approx}11.48$) yields the highest PSNR (26.55\,dB) and SSIM (0.7579), but at the cost of notably degraded perceptual metrics across the board. This suggests that an overly strict constraint leaves too little room for the semantic LoRA to learn meaningful perceptual improvements, effectively over-preserving the pixel-level subspace. This controlled trade-off validates the OGP design: the orthogonal projection effectively decouples the two learning objectives, and $\tau{=}0.95$ provides the best overall perception-fidelity balance, which we adopt as the default setting.

\subsection{Effect of Number of Experts and Top-$k$}
\label{sec:abl_experts_topk}

We study the impact of the number of LoRA experts $N$ and the top-$k$ routing parameter on FreqMoE performance. Table~\ref{tab:abl_experts} evaluates different configurations on the RealSR dataset. Our default setting uses $N{=}4$ experts with top-$k{=}2$ routing.

\begin{table}[t]
\centering
\caption{Ablation study on the number of experts ($N$) and top-$k$ routing in FreqMoE on the RealSR dataset. The best results are highlighted in \textcolor{best}{red} and the second best in \textcolor{second}{blue}.}
\label{tab:abl_experts}
\setlength{\tabcolsep}{2pt}
\scalebox{0.78}{
\begin{tabular}{ccccccccccc}
\toprule
$N$ & Top-$k$ & PSNR$\uparrow$ & SSIM$\uparrow$ & LPIPS$\downarrow$ & DISTS$\downarrow$ & FID$\downarrow$ & NIQE$\downarrow$ & CLIPIQA$\uparrow$ & MUSIQ$\uparrow$ & MANIQA$\uparrow$ \\
\midrule
1 & 1 & 26.20 & 0.7432 & 0.2651 & 0.2021 & 113.39 & \textcolor{second}{5.44} & \textcolor{second}{0.6646} & \textcolor{second}{69.13} & \textcolor{second}{0.6557} \\
2 & 1 & 26.42 & 0.7516 & 0.2594 & 0.2005 & 115.67 & 5.49 & 0.6543 & 68.89 & 0.6481 \\
2 & 2 & 26.44 & 0.7548 & 0.2591 & 0.2021 & 113.23 & 5.52 & 0.6470 & 68.72 & 0.6457 \\
\rowcolor{gray!10}
4 & 2 & 26.27 & 0.7481 & \textcolor{second}{0.2537} & \textcolor{second}{0.1951} & \textcolor{best}{108.91} & \textcolor{best}{5.32} & 0.6630 & \textcolor{best}{69.39} & \textcolor{best}{0.6586} \\
4 & 4 & 26.31 & 0.7523 & 0.2549 & 0.1963 & 109.99 & 5.49 & 0.6549 & 68.73 & 0.6513 \\
6 & 2 & 26.35 & 0.7550 & 0.2550 & 0.1972 & 116.26 & 5.68 & 0.6607 & 68.60 & 0.6505 \\
6 & 3 & 26.33 & 0.7538 & 0.2586 & 0.2001 & 119.30 & 5.67 & 0.6630 & 68.64 & 0.6534 \\
8 & 2 & \textcolor{best}{26.56} & \textcolor{second}{0.7575} & \textcolor{best}{0.2508} & \textcolor{best}{0.1948} & \textcolor{second}{109.70} & \textcolor{second}{5.44} & 0.6512 & 68.25 & 0.6454 \\
8 & 4 & \textcolor{second}{26.53} & \textcolor{best}{0.7584} & 0.2585 & 0.1998 & 117.34 & 5.51 & \textcolor{best}{0.6650} & \textcolor{second}{68.77} & 0.6466 \\
\bottomrule
\end{tabular}}
\end{table}

The single-expert baseline ($N{=}1$) uses only OGP without MoE routing. Increasing $N$ generally improves fidelity (e.g., $N{=}8$, top-$k{=}2$ achieves the best PSNR, LPIPS, and DISTS), but perceptual quality does not scale accordingly. $N{=}4$, top-$k{=}2$ yields the best perceptual metrics (FID 108.91, NIQE 5.32, MANIQA 0.6586) while still improving over the baseline by $+0.07$\,dB PSNR and $-4.48$ FID. For top-$k$, activating more experts generally hurts perceptual quality ($N{=}6$: $k{=}2$ vs.\ $k{=}3$; $N{=}8$: $k{=}2$ vs.\ $k{=}4$), indicating that sparse routing encourages better expert specialization. We adopt $N{=}4$, top-$k{=}2$ for the best perceptual-fidelity balance with lower computational cost.

\subsection{Effect of FreqMoE on Expert Routing}
\label{sec:abl_mole_routing}

\begin{figure}[t]
\centering
\includegraphics[width=1.\linewidth]{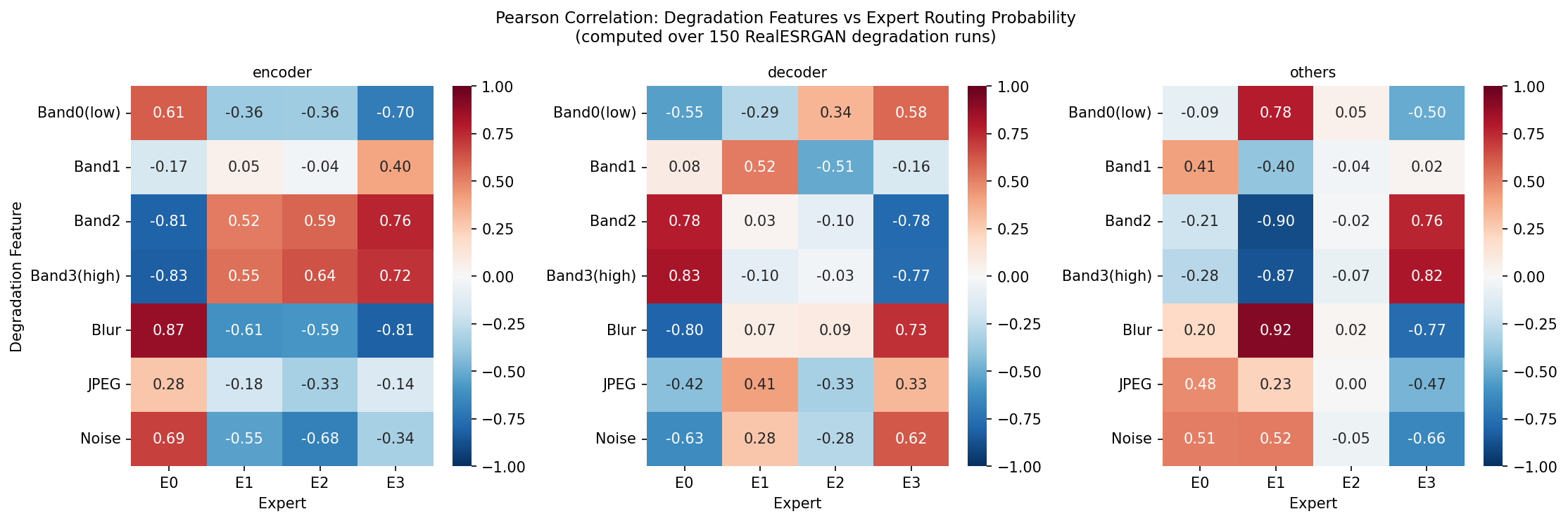}
\caption{Pearson correlation between degradation features and expert routing probabilities in FreqMoE, computed over 150 RealESRGAN degradation runs. Each heatmap corresponds to a different UNet component (encoder, decoder, others). Rows represent degradation features: radial frequency band energies (Band0--Band3, from low to high frequency) and scalar degradation indicators (Blur, JPEG, Noise). Columns represent the four experts (E0--E3). Distinct correlation patterns across experts confirm that frequency-guided routing enables meaningful degradation-aware specialization.}
\label{fig:expert_routing}
\end{figure}

To further validate that FreqMoE learns interpretable expert specialization, we compute the Pearson correlation between the degradation feature vector $\mathbf{f}$ (Sec.~3.2 of the main paper) and the routing probabilities assigned to each expert across 150 images generated by the RealESRGAN degradation pipeline. Fig.~\ref{fig:expert_routing} shows the resulting correlation heatmaps for the encoder, decoder, and other (mid-block and skip connections) layers of the UNet.

Several observations emerge. First, experts exhibit distinct correlation profiles with respect to different degradation features, confirming that the frequency-guided gating network learns to specialize experts rather than routing uniformly. Second, the correlation patterns vary across UNet components: encoder layers tend to show stronger differentiation among experts for high-frequency features (Band3, Noise), while decoder layers exhibit more pronounced specialization for structural degradations (Blur, JPEG). This is consistent with the hierarchical nature of UNet, where encoder layers process increasingly abstract features and decoder layers reconstruct spatial details. Third, no single expert dominates across all degradation types, indicating that the load balancing loss effectively prevents expert collapse and that the top-$k{=}2$ routing allows complementary experts to collaborate on mixed-degradation inputs.

\begin{sidewaysfigure}
\centering
\includegraphics[width=\linewidth]{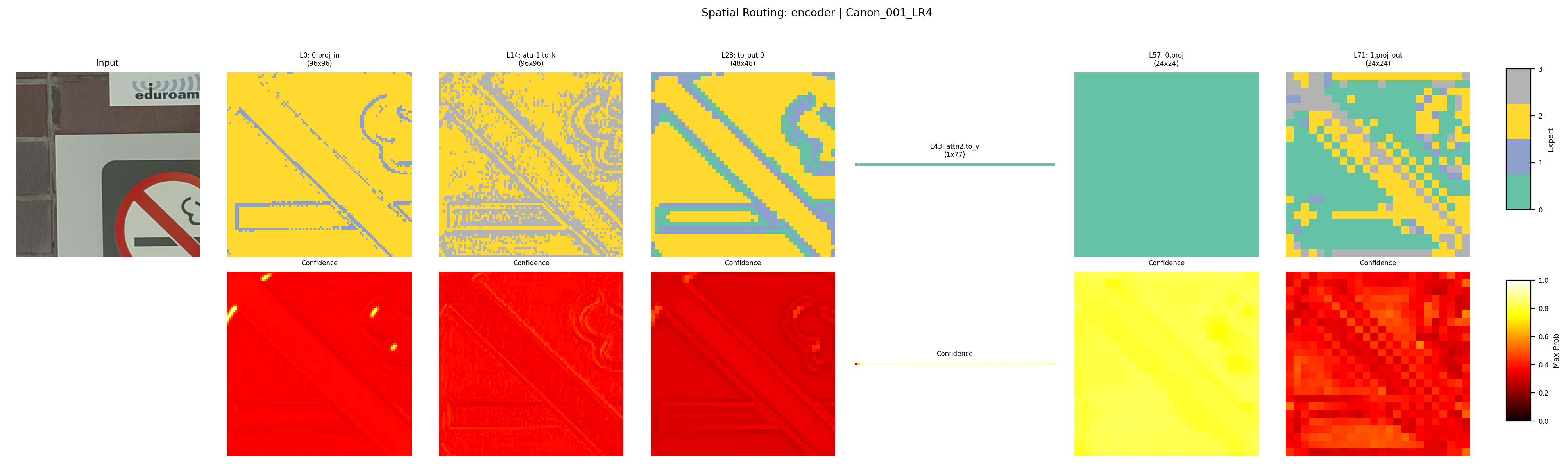}
\caption{Spatial routing visualization across encoder layers for a real-world image from the RealSR dataset. Top row: expert assignment maps at each encoder layer, where colors indicate the selected top-$k$ experts at each spatial location. Bottom row: corresponding routing confidence maps. The routing patterns reveal that FreqMoE adapts expert selection spatially: structurally complex regions (e.g., text, sign edges) and smooth background regions activate different expert combinations. Moreover, the routing evolves across layers: early encoder layers exhibit fine-grained spatial differentiation, while deeper layers show coarser, more semantically coherent routing. This per-image spatial view complements the aggregate correlation analysis in Fig.~\ref{fig:expert_routing}, confirming that FreqMoE achieves both image-level degradation adaptation and spatially-aware expert specialization within each image.}
\label{fig:spatial_routing}
\end{sidewaysfigure}

To complement the above aggregate analysis, Fig.~\ref{fig:spatial_routing} visualizes the spatial routing decisions across encoder layers for a single real-world image. The visualization confirms that FreqMoE not only adapts routing based on global degradation characteristics, but also differentiates expert assignments spatially within each image, assigning different expert combinations to structurally distinct regions such as text, edges, and smooth backgrounds.

\subsection{Effect of Number of Frequency Bands ($K$)}
\label{sec:abl_freq_bands}

The number of radial frequency bands $K$ determines the granularity at which the degradation feature extractor captures the spectral energy distribution (Sec.~3.2 of the main paper). Since the degradation feature vector has dimension $d_f = K + 3$ (frequency band energies plus three scalar degradation scores), $K$ also affects the input dimensionality of the frequency-modulated gating network. Our default setting uses $K{=}4$, motivated by the four principal degradation types in the RealESRGAN pipeline~\cite{wang2021real}: blur (low-frequency suppression), resize (mid-frequency aliasing), noise (high-frequency elevation), and JPEG compression (block artifacts). Table~\ref{tab:abl_freq_bands} evaluates $K \in \{2, 4, 8\}$ on the RealSR dataset.

\begin{table}[t]
\centering
\caption{Ablation study on the number of frequency bands $K$ in the degradation feature extractor on the RealSR dataset. The feature dimension is $d_f = K + 3$. The best results are highlighted in \textcolor{best}{red} and the second best in \textcolor{second}{blue}.}
\label{tab:abl_freq_bands}
\setlength{\tabcolsep}{2pt}
\scalebox{0.78}{
\begin{tabular}{ccccccccccc}
\toprule
$K$ & $d_f$ & PSNR$\uparrow$ & SSIM$\uparrow$ & LPIPS$\downarrow$ & DISTS$\downarrow$ & FID$\downarrow$ & NIQE$\downarrow$ & CLIPIQA$\uparrow$ & MUSIQ$\uparrow$ & MANIQA$\uparrow$ \\
\midrule
2 & 5 & \textcolor{best}{26.48} & \textcolor{best}{0.7561} & \textcolor{best}{0.2528} & \textcolor{second}{0.1968} & \textcolor{second}{113.18} & 5.75 & 0.6407 & 68.28 & 0.6454 \\
\rowcolor{gray!10}
4 & 7 & 26.27 & 0.7481 & \textcolor{second}{0.2537} & \textcolor{best}{0.1951} & \textcolor{best}{108.91} & \textcolor{best}{5.32} & \textcolor{best}{0.6630} & \textcolor{best}{69.39} & \textcolor{best}{0.6586} \\
8 & 11 & \textcolor{second}{26.44} & \textcolor{second}{0.7542} & 0.2582 & 0.1994 & 113.94 & \textcolor{second}{5.46} & \textcolor{second}{0.6526} & \textcolor{second}{68.76} & \textcolor{second}{0.6498} \\
\bottomrule
\end{tabular}}
\end{table}

$K{=}2$ achieves the highest fidelity scores (PSNR $26.48$\,dB, SSIM $0.7561$, LPIPS $0.2528$), yet its coarse two-band decomposition (low vs.\ high frequency) cannot distinguish degradation types that occupy overlapping spectral regions (e.g., blur vs.\ resize aliasing in low-mid frequencies), leading to the worst perceptual quality across all no-reference metrics. $K{=}8$ provides finer spectral resolution and also improves fidelity over $K{=}4$ (PSNR $+0.17$\,dB), but the highly correlated adjacent bands introduce redundancy into the gating signal, biasing routing toward pixel-level optimization at the expense of expert specialization for perceptual quality: FID increases by $5.03$, NIQE by $0.14$, and CLIPIQA, MUSIQ, MANIQA all drop compared to $K{=}4$. $K{=}4$ achieves the best overall fidelity-perception balance, aligning with the intuition that four bands naturally correspond to the four principal degradation categories in the RealESRGAN pipeline: blur, resize, noise, and JPEG compression.

\subsection{Effect of OGP Projection Target}
\label{sec:abl_ogp_target}

By default, OGP projects gradients only on the semantic LoRA $B$ matrices (up-projection, $B_{sem} \in \mathbb{R}^{d_{out} \times r}$). We first provide a theoretical justification for this choice, then validate it empirically.

\noindent\textbf{Theoretical justification.}
The goal of OGP is to ensure that the effective weight change of the semantic LoRA, $\Delta W_{sem} = B_{sem} A_{sem}$, has no component in the pixel-level subspace spanned by $U_{\tilde{k}}$, i.e., $U_{\tilde{k}}^T \Delta W_{sem} = \mathbf{0}$. We show that constraining only $B_{sem}$ is \emph{sufficient} to achieve this. Expanding the orthogonality condition:
\begin{equation}
  U_{\tilde{k}}^T \Delta W_{sem} = U_{\tilde{k}}^T (B_{sem} A_{sem}) = (U_{\tilde{k}}^T B_{sem})\, A_{sem}.
  \label{eq:b_sufficient}
\end{equation}
If OGP ensures that each column of $B_{sem}$ lies in the null space of $U_{\tilde{k}}^T$, i.e., $U_{\tilde{k}}^T B_{sem} = \mathbf{0}$, then Eq.~\eqref{eq:b_sufficient} yields $U_{\tilde{k}}^T \Delta W_{sem} = \mathbf{0} \cdot A_{sem} = \mathbf{0}$, \emph{regardless of} $A_{sem}$. Therefore, projecting $B_{sem}$ gradients alone provides a \emph{complete guarantee} that the full semantic weight change $\Delta W_{sem}$ is orthogonal to the pixel-level subspace for any $A_{sem}$.

Conversely, projecting only $A_{sem}$ gradients using the right singular vectors $V_{\tilde{k}}$ does \emph{not} provide the same guarantee. Even if $A_{sem}$ lies in the null space of $V_{\tilde{k}}^T$, the product $U_{\tilde{k}}^T B_{sem} A_{sem}$ can still be nonzero whenever $B_{sem}$ has components along $U_{\tilde{k}}$. Hence, $B$-only projection is both necessary and sufficient for output-space orthogonality, while $A$-only projection is neither.

\noindent\textbf{Empirical validation.}
One might still ask whether additionally projecting $A$ matrix gradients onto the null space of $V_{\tilde{k}}$ could yield further benefits. We compare both strategies in Table~\ref{tab:abl_ogp_target}.

\begin{table}[t]
\centering
\setlength{\extrarowheight}{0pt}
\addtolength{\extrarowheight}{\aboverulesep}
\addtolength{\extrarowheight}{\belowrulesep}
\setlength{\aboverulesep}{0pt}
\setlength{\belowrulesep}{0pt}
\caption{Ablation study on OGP projection target on RealSR. ``$B$ only'' (default) projects semantic gradients onto the null space of $U_{\tilde{k}}$ for $B$ matrices. ``Both $A$ and $B$'' additionally projects $A$ matrix gradients onto the null space of $V_{\tilde{k}}$. The best results are highlighted in \textcolor{best}{red}.}
\label{tab:abl_ogp_target}
\setlength{\tabcolsep}{1mm}
\scalebox{0.78}{
\begin{tabular}{lccccccccc}
\toprule
OGP Target & PSNR$\uparrow$ & SSIM$\uparrow$ & LPIPS$\downarrow$ & DISTS$\downarrow$ & FID$\downarrow$ & NIQE$\downarrow$ & CLIPIQA$\uparrow$ & MUSIQ$\uparrow$ & MANIQA$\uparrow$ \\
\midrule
Both $A$ and $B$ & \textcolor{best}{26.43} & \textcolor{best}{0.7527} & 0.2611 & 0.1975 & 115.89 & 5.59 & 0.6616 & 68.35 & 0.6456 \\
$B$ only (Ours) & 26.27 & 0.7481 & \textcolor{best}{0.2537} & \textcolor{best}{0.1951} & \textcolor{best}{108.91} & \textcolor{best}{5.32} & \textcolor{best}{0.6630} & \textcolor{best}{69.39} & \textcolor{best}{0.6586} \\
\bottomrule
\end{tabular}}
\end{table}

Projecting both $A$ and $B$ improves pixel-fidelity metrics (PSNR: $+0.16$\,dB, SSIM: $+0.005$). However, this comes at the cost of degraded perceptual quality: LPIPS worsens by $0.007$, DISTS increases by $0.002$, and FID degrades notably ($+6.98$). No-reference metrics (MUSIQ, MANIQA) also drop.

These results are consistent with the theoretical analysis above. Projecting $B$ alone already provides a \emph{complete} orthogonality guarantee (Eq.~\eqref{eq:b_sufficient}), so the additional $A$-projection does not further reduce subspace interference but instead over-constrains the semantic LoRA's learning capacity. By restricting both the output \emph{and} input subspaces, the ``Both $A$ and $B$'' variant reduces the effective rank of $\Delta W_{sem}$, pushing the model toward higher fidelity at the expense of perceptual enhancement. The $B$-only strategy preserves the semantic branch's freedom to explore complementary perceptual features in the input space while fully preventing overlap in the output space.

\subsection{Effect of Applying FreqMoE to Both Branches}
\label{sec:abl_freqmoe_both}

We investigate whether applying the FreqMoE module to both the pixel and semantic branches improves performance, compared to our default design that uses FreqMoE only for the pixel branch and a standard LoRA for the semantic branch. Table~\ref{tab:abl_freqmoe_both} presents the comparison on RealSR.

\begin{table}[t]
\centering
\setlength{\extrarowheight}{0pt}
\addtolength{\extrarowheight}{\aboverulesep}
\addtolength{\extrarowheight}{\belowrulesep}
\setlength{\aboverulesep}{0pt}
\setlength{\belowrulesep}{0pt}
\caption{Ablation study on applying FreqMoE to both branches vs.\ the default design (FreqMoE for pixel branch only) on RealSR\@. Applying FreqMoE to both branches improves fidelity metrics (PSNR, SSIM) but degrades perceptual and no-reference quality metrics. The best results are highlighted in \textcolor{best}{red}.}
\label{tab:abl_freqmoe_both}
\setlength{\tabcolsep}{1mm}
\scalebox{0.78}{
\begin{tabular}{lccccccccc}
\toprule
Configuration & PSNR$\uparrow$ & SSIM$\uparrow$ & LPIPS$\downarrow$ & DISTS$\downarrow$ & FID$\downarrow$ & NIQE$\downarrow$ & CLIPIQA$\uparrow$ & MUSIQ$\uparrow$ & MANIQA$\uparrow$ \\
\midrule
Both & \textcolor{best}{26.47} & \textcolor{best}{0.7567} & 0.2583 & 0.2008 & 115.67 & 5.47 & 0.6594 & 68.43 & 0.6484 \\
Pixel only (Ours) & 26.27 & 0.7481 & \textcolor{best}{0.2537} & \textcolor{best}{0.1951} & \textcolor{best}{108.91} & \textcolor{best}{5.32} & \textcolor{best}{0.6630} & \textcolor{best}{69.39} & \textcolor{best}{0.6586} \\
\bottomrule
\end{tabular}}
\end{table}

Applying FreqMoE to both branches improves pixel-fidelity metrics (PSNR: $+0.20$\,dB, SSIM: $+0.009$). However, this comes at the cost of degraded perceptual and no-reference quality: LPIPS increases by $0.005$, CLIPIQA drops by $0.004$, MUSIQ drops by $1.0$, and MANIQA drops by $0.010$.

This result supports our architectural choice. The semantic branch is designed to learn complementary perceptual features that enhance visual quality beyond pixel-level fidelity. Applying frequency-guided routing to the semantic branch biases it toward the same degradation-aware, pixel-oriented optimization as the pixel branch, effectively constraining its capacity for semantic enhancement. In contrast, using a standard LoRA for the semantic branch, combined with OGP to ensure orthogonality, allows it to freely explore the perceptual quality manifold without being anchored to frequency-domain degradation patterns.

\section{More Qualitative Results}
\label{sec:more_qualitative}

We provide additional visual comparisons of FreqOrtho-SR with state-of-the-art methods.

\begin{figure}[!ht]
\centering
\includegraphics[width=1.\linewidth]{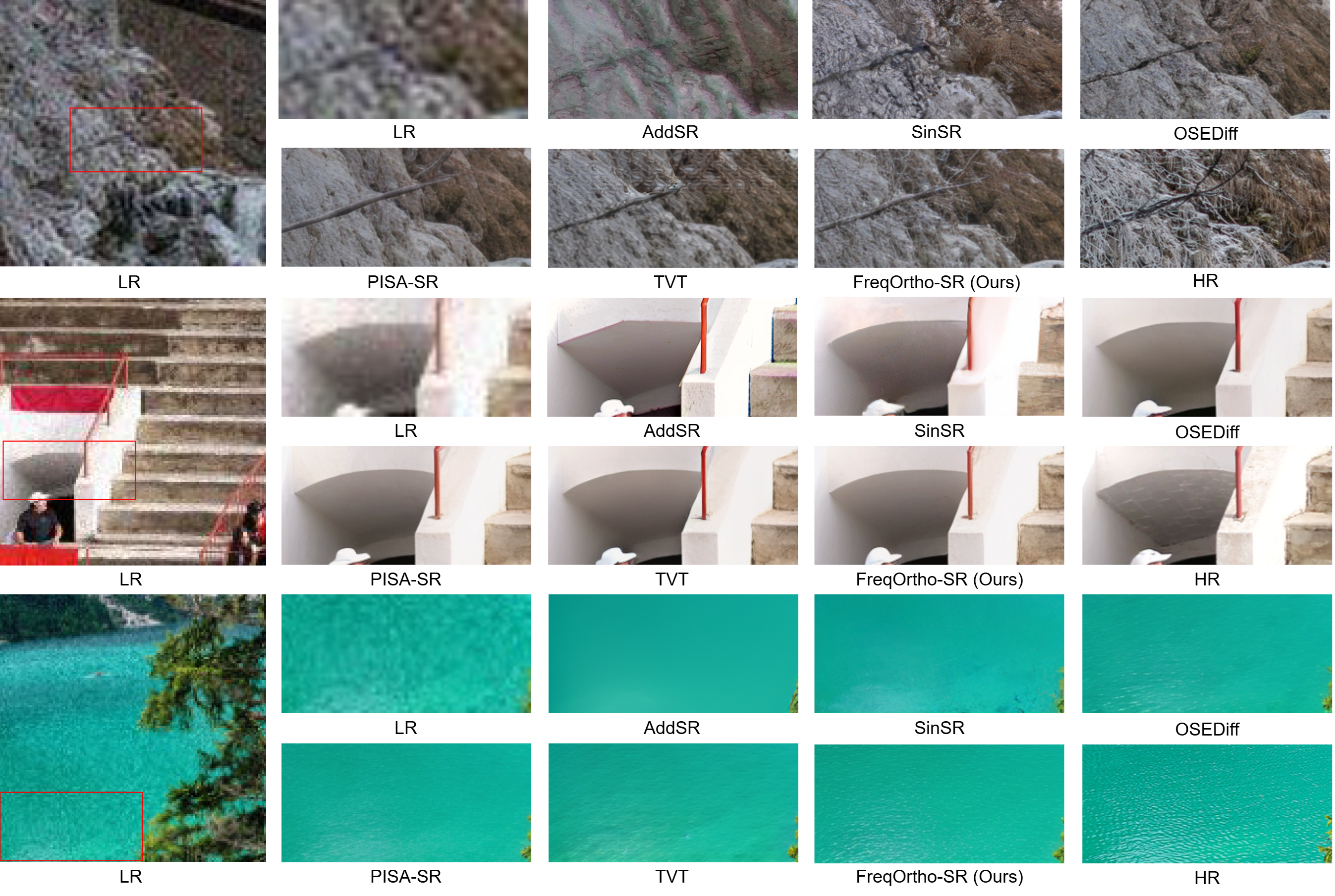}
\caption{Additional visual comparisons between FreqOrtho-SR and competing methods from the main paper. Please zoom in for a better view.}
\label{fig:more_qual_additional}
\end{figure}

\begin{figure}[p]
\centering
\includegraphics[width=1.\linewidth]{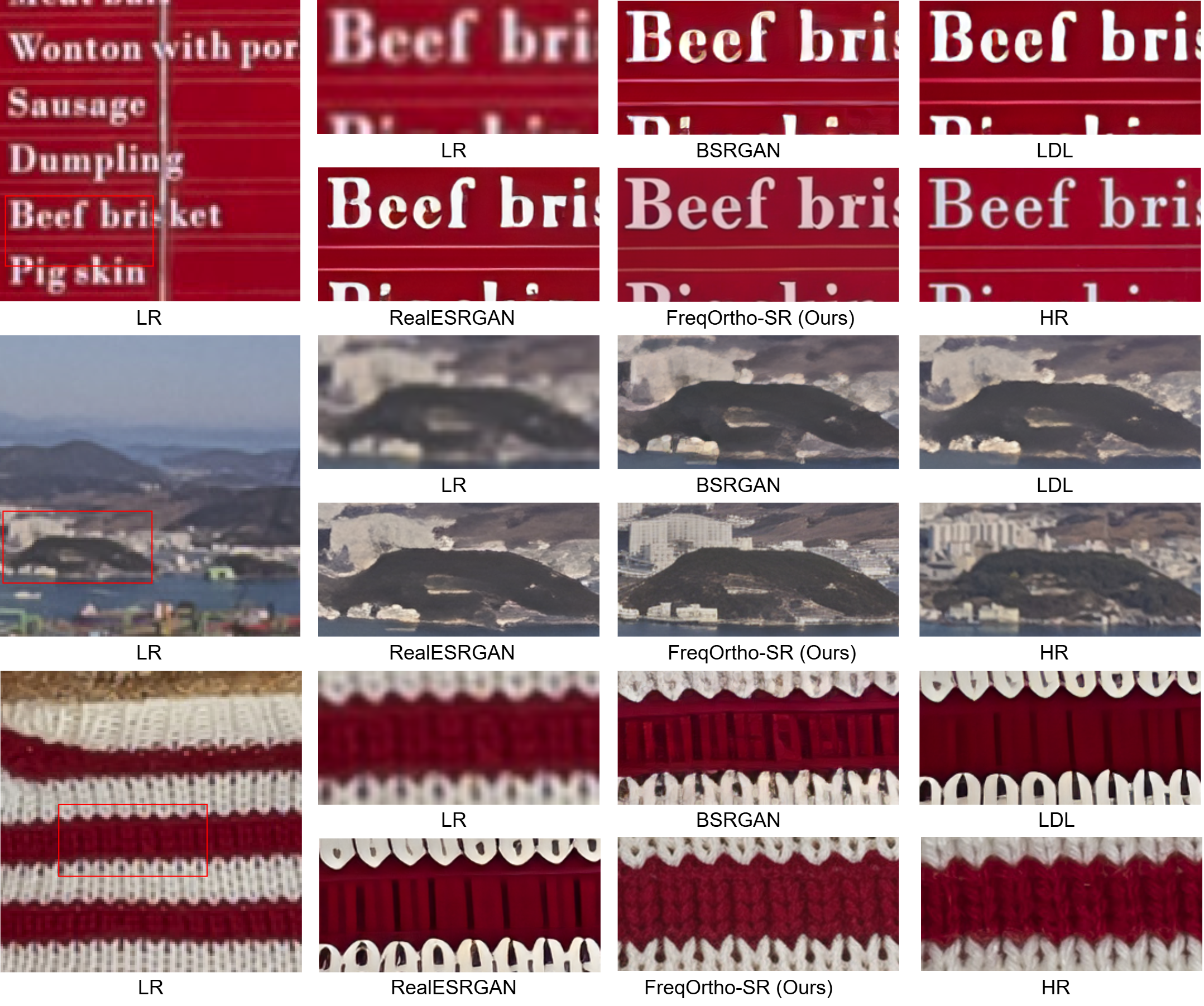}
\caption{Visual comparisons between FreqOrtho-SR and GAN-based SR methods. Please zoom in for a better view.}
\label{fig:gan_qual}
\end{figure}

\begin{figure}[p]
\centering
\includegraphics[width=1.\linewidth]{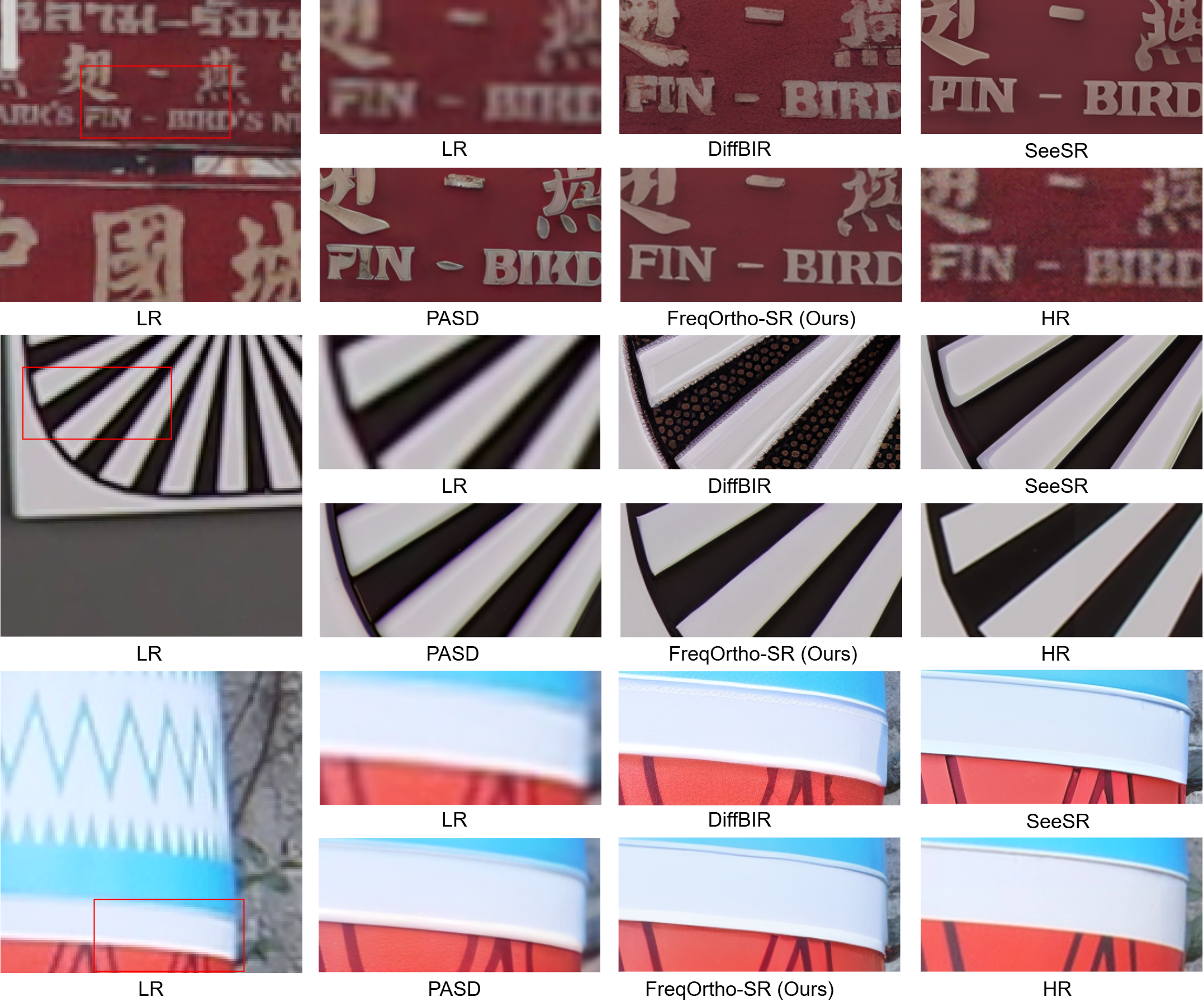}
\caption{Visual comparisons between FreqOrtho-SR and multi-step SR methods. Please zoom in for a better view.}
\label{fig:multistep_qual}
\end{figure}

As shown in Fig.~\ref{fig:more_qual_additional}, FreqOrtho-SR consistently produces sharper and more detailed outputs compared to other one-step DM-based methods, which tend to suffer from blurriness or hallucinated textures. Fig.~\ref{fig:gan_qual} and Fig.~\ref{fig:multistep_qual} further compare with GAN-based and multi-step DM-based methods, respectively, where similar observations hold: competing methods either over-smooth fine details or introduce artifacts that deviate from the ground truth, while FreqOrtho-SR maintains faithful and high-quality reconstructions across all scenes.

\clearpage

\clearpage

\section{Limitations and Future Works}
\label{sec:limitations}

\noindent\textbf{Limitations.}
While FreqOrtho-SR achieves strong results on the perception-fidelity trade-off, its current OGP design relies on a fixed SVD basis extracted once between the two training phases. In our setting, this basis remains valid during Phase~2 because the pixel-level FreqMoE is frozen, so the pixel subspace does not drift. This fixed-basis design may be insufficient in future settings where pixel- and semantic-level branches are updated jointly or co-evolve during training.

\noindent\textbf{Future works.}
Our results demonstrate that orthogonal gradient projection is a promising principle for mitigating subspace interference between pixel-fidelity and semantic-enhancement objectives in real-world SR. While null-space projection has been extensively studied in continual learning~\cite{saha2021gradient, zeng2019continual, lopez2017gradient}, its application to multi-objective image super-resolution remains largely unexplored, and we believe there is significant room for further investigation. For future co-evolving branch settings, more advanced techniques could be explored to make the projection adaptive, task-aware, and geometrically principled. Such extensions could further improve the pixel-semantic decoupling and push the perception-fidelity Pareto frontier in real-world image super-resolution.

\clearpage
\bibliographystyle{splncs04}
\bibliography{main}